\pdfoutput=1
\PassOptionsToPackage{colorlinks=true, urlcolor=blue, linkcolor=blue}{hyperref}

\documentclass{article}

\usepackage{arxiv}

\usepackage[utf8]{inputenc}
\usepackage[T1]{fontenc}

\usepackage{amsmath}
\usepackage{amsthm}
\usepackage{amssymb}
\usepackage{amsfonts}
\usepackage{mathtools}
\usepackage{bm}

\usepackage{graphicx}
\usepackage{subcaption}
\usepackage{wrapfig}
\usepackage{booktabs}
\usepackage{float}

\usepackage[ruled,vlined]{algorithm2e}

\usepackage[table]{xcolor} %
\usepackage[most]{tcolorbox}

\usepackage{multirow}
\usepackage{threeparttable}
\usepackage{tabularx}
\usepackage{microtype}
\usepackage{nicefrac}
\usepackage{moresize}
\usepackage{comment}
\usepackage{csquotes}
\usepackage{cancel}
\usepackage{lipsum}

\usepackage{enumitem}
\usepackage{natbib}
\usepackage{doi}

\usepackage{fontawesome5}

\usepackage[capitalize,noabbrev]{cleveref}
\usepackage{url}

\usepackage{subfiles}

\usepackage{footnote}
\makesavenoteenv{tabular}

\usepackage[textsize=tiny]{todonotes}

\usepackage{listings}
\usepackage{parskip}

\definecolor{catcorrect}{HTML}{1A7F4B}\definecolor{catcorrectbg}{HTML}{E6F4EC}
\definecolor{catminor}{HTML}{B86400}\definecolor{catminorbg}{HTML}{FFF3CD}
\definecolor{catfixable}{HTML}{1565C0}\definecolor{catfixablebg}{HTML}{E3F2FD}
\definecolor{catfatal}{HTML}{C62828}\definecolor{catfatalbg}{HTML}{FDECEA}

\DeclareMathOperator*{\argmax}{argmax}

\title{\benchname: Benchmarking LLMs’ Reasoning for Differential Privacy}

\date{}

\author{
\parbox{0.95\textwidth}{\centering\normalfont
\textbf{Erchi Wang}$^{1\dagger}$ \hspace{0.4em}
\textbf{Pengrun Huang}$^{2\dagger}$ \hspace{0.4em}
\textbf{Eli Chien}$^{3}$ \hspace{0.4em}
\textbf{Om Thakkar}$^{4}$ \hspace{0.4em}
\textbf{Kamalika Chaudhuri}$^{2}$ \hspace{0.4em}
\textbf{Yu-Xiang Wang}$^{1}$\\[2pt]
\textbf{Ruihan Wu}$^{4\dagger,*}$\\[6pt]
$^{1}$ Halıcıoğlu Data Science Institute, UC San Diego\\
$^{2}$ Department of Computer Science and Engineering, UC San Diego\\
$^{3}$ Department of Electrical Engineering, National Taiwan University\\
$^{4}$ OpenAI\\
$^\dagger$ Denotes core contribution, $^*$ Work performed at UC San Diego\\
Correspondence to: \texttt{erw011@ucsd.edu}, \texttt{ruihan@openai.com}
}
}

\renewcommand{\shorttitle}{\benchname: Benchmarking LLMs’ Reasoning for Differential Privacy}

\definecolor{purple}{HTML}{7C3AED}
\definecolor{amber}{HTML}{F59E0B}
\definecolor{teal}{HTML}{0D9488}

\theoremstyle{plain}
\newtheorem{theorem}{Theorem}[section]

\theoremstyle{definition}
\newtheorem{definition}[theorem]{Definition}

\theoremstyle{remark}

\newcommand{\gptmin}{GPT-5-Minimal}
\newcommand{\gpthigh}{GPT-5-High}
\newcommand{\gptfive}{GPT-5}
\newcommand{\gptffh}{GPT-5.4-High}
\newcommand{\gptfvh}{GPT-5.5-High}
\newcommand{\judgemodel}{GPT-5.4-High}

\newcommand{\gemhigh}{Gemini-3-Pro}
\newcommand{\gemthro}{Gemini-3.1-Pro}
\newcommand{\gemmin}{Gemini-2.5-Flash}

\newcommand{\claudesonnet}{Claude-Sonnet-4.5}
\newcommand{\claudeopus}{Claude-Opus-4.5}

\newcommand{\qweninst}{Qwen3-30-Instruct}
\newcommand{\qwenthink}{Qwen3-30-Think}

\newcommand{\dpskrone}{DeepSeek-R1}
\newcommand{\dpskchat}{DeepSeek-V3.1-chat}

\newcommand{\goeprove}{Goedel-Prover-V2}

\newcommand{\catone}{Category 1}
\newcommand{\catwo}{Category 2}
\newcommand{\benchname}{DPrivBench}

\begin{document}
\maketitle

\begin{abstract}

Differential privacy (DP) has a wide range of applications for protecting data privacy, but designing and verifying DP algorithms requires expert-level reasoning, creating a high barrier for non-expert practitioners. Prior works either rely on specialized verification languages that demand substantial domain expertise or remain semi-automated and require human-in-the-loop guidance. In this work, we investigate whether large language models (LLMs) can automate DP reasoning.
We introduce \benchname{}, a benchmark in which each instance asks whether a function or algorithm satisfies a stated DP guarantee under specified assumptions. 
The benchmark is carefully designed to cover a broad range of DP topics, span diverse difficulty levels, and resist shortcut reasoning through trivial pattern matching.
Experiments show that while the strongest models handle textbook mechanisms well, all models struggle with advanced algorithms, revealing substantial gaps in current DP reasoning capabilities. 
Through further analytic study and failure-mode analysis, we identify several promising directions for improving automated DP reasoning. 
Our benchmark provides a solid foundation for developing and evaluating such methods, and complements existing benchmarks for mathematical reasoning.
\end{abstract}

\vspace{0.5em}
\noindent\textbf{Resources: }
\href{https://github.com/erchiw/DPriv-Bench.git}{\faGithub\ Code}
\quad | \quad
\href{https://huggingface.co/datasets/erchiw/DPriv-Bench}{%
  \includegraphics[height=2.5ex]{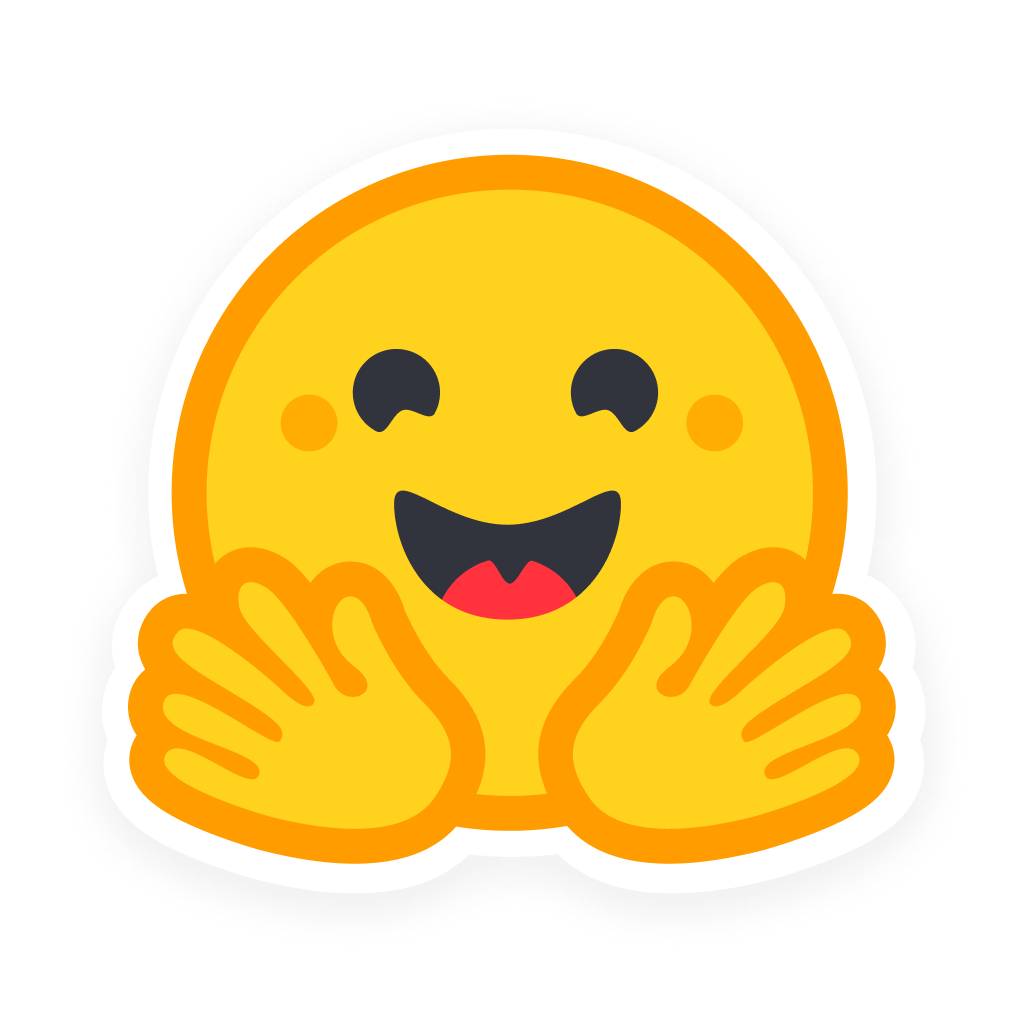}\ Dataset
}

\section{Introduction}

Differential privacy (DP)~\citep{dwork2006calibrating} has emerged as the gold standard for data privacy, offering rigorous mathematical guarantees that protect individual information while still enabling meaningful analyses. Its impact spans a wide range of applications from national statistics released by government agencies~\citep{abowd2018us, nist2020} to the handling of user data by major technology companies~\citep{google_dp_blog_2021, meta_pets_adtech_2025, icml2021_virtual_11631, apple_dp_aggregate_trends}.

Despite its broad applicability, developing and deploying DP mechanisms for specific use cases often requires substantial expertise. Designing an algorithm with a target privacy budget involves careful reasoning with specialized knowledge in DP, an error-prone task even for DP researchers \citep[see, e.g.,][]{lyu2017understanding}.  %
This high barrier prevents non-experts from utilizing DP in their application.  %

Towards the long-term goal of developing and deploying DP algorithms automatically, 
the literature has primarily advanced along two directions.
\textit{Programmatic DP verification}~\citep{reed2010distance, barthe2014proving, barthe2016proving, albarghouthi2017synthesizing, sato2019approximate} formally verify DP guarantees by checking symbolic proofs or synthesizing mechanisms from formal algorithm specifications. 
While these systems provide strong soundness guarantees, they typically require substantial domain expertise to encode algorithms in specialized verification languages, which limits their accessibility to non-expert users.
Another complementary line of work can be viewed as \emph{semi-automated DP}, pioneered by DPCheatSheet~\citep{chu2025dpcheatsheet}, in which LLMs are used to help non-experts design and implement DP algorithms interactively.

With the rapid progress of large language models (LLMs), especially their strong performance on general mathematical reasoning tasks~\citep{huang2025winning, openai_gpt5_announcement, deepmind_gemini3pro_modelcard}, it is natural to ask whether they can assist with reasoning about differential privacy (DP), for example, by identifying flaws in DP proofs or verifying the privacy guarantees of stated DP algorithms. Unlike approaches based on formal verification languages or intensive human-in-the-loop guidance, this direction treats the LLM as the primary reasoning agent: given an algorithm description in natural language or \LaTeX, as is often the case in practice, the LLM is asked to reason about whether the algorithm satisfies a claimed DP guarantee. A sufficiently capable agent could support both beginners in developing basic DP algorithms and experts by serving as a complementary “second check” on the correctness of system design.

This paper focuses on this emerging direction as the first work: 
\begin{center}
	\textit{Can LLM reason about the DP guarantees of algorithms?}
\end{center}

To study this question, we introduce \textbf{\benchname{}}, a benchmark of curated instances that describe an algorithm and its assumptions, and ask LLMs to verify the stated DP guarantee.  The benchmark is designed according to three guiding principles for comprehensiveness and reliability: broad topic coverage, diverse difficulty levels, and resistance to shortcut reasoning through trivial pattern matching.The benchmark consists of two complementary categories. \benchname{} contains two complementary categories: \catone{} covers textbook sensitivity-based mechanisms, while \catwo{} targets advanced DP algorithms requiring substantially more sophisticated reasoning.

\begin{figure}[t]
\includegraphics[width=\linewidth]{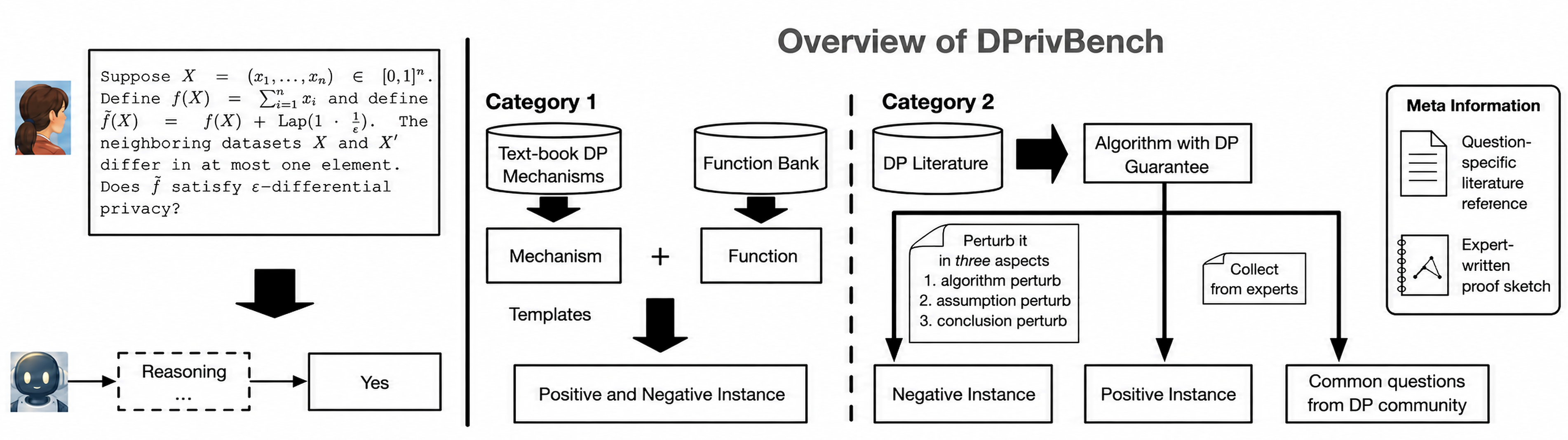}
\caption{Overview of \benchname{}. The left panel shows a representative reasoning instance, while the right panel summarizes the benchmark construction: \catone{} mechanism-level instances from a function bank and \catwo{} algorithm-level instances from the DP literature. Overall, \benchname{} contains 588 \catone{} and 125 \catwo{} instances, totaling 713 instances. }
\label{fig:overview}
\end{figure}

By evaluating a diverse set of state-of-the-art language models on \benchname{}, we obtain several key observations. For \textit{foundational sensitivity-based DP mechanisms}, the strong closed-source models (eg. \gpthigh{} and \gemhigh{}) equipped with an enhanced reasoning mode achieve high accuracy, while all other models exhibit non-negligible error rates. For \textit{advanced differential privacy algorithms with nontrivial analysis}, no evaluated model demonstrates consistently strong performance, even among stronger models such as \gptfvh{} and \gemthro{}. These results suggest that while current models are sufficient for textbook-level DP reasoning and may serve as useful aids for beginners, a gap remains for reliably analyzing modern DP algorithms. 

We conduct further study supported by our collected metadata, aiming to guide the future study for better LLM reasoning.
\textit{First}, we evaluate whether providing explicit references that mimic information retrieval from external sources improves accuracy. In this case, performance increases noticeably, pointing to a promising direction for future tools that integrate LLM reasoning with curated DP knowledge bases.
\textit{Second}, we further utilize LLM-as-Judge to perform fine-grind analysis of response, and find that the prediction produced by strong reasoning models is trustworthy in general.
\textit{Last}, We conduct targeted case studies to characterize common model failure modes, highlighting key aspects to address when improving reasoning trajectories.

\textbf{Our benchmark serves as a cornerstone for advancing the automation of DP reasoning with LLMs}, paving the way for broader deployment of DP by reducing expertise requirements and enhancing efficiency for experts. By guiding the development of stronger reasoning agents, our benchmark can further promote the wider adoption of privacy-preserving techniques in real-world applications.
Beyond its practical value for privacy research, \textbf{it also serves as a new and challenging testbed for mathematical reasoning.} Since DP is typically taught as a graduate-level topic in applied mathematics and theoretical computer science, our benchmark complements existing math reasoning datasets (e.g., GSM8K~\citep{cobbe2021gsm8k}, GPQA~\citep{rein2024gpqa}, MATH-Perturb~\citep{huang2025mathperturb}). 

\newpage

\section{Problem Set-Up}
\label{sec:problem_setup}
\textbf{Differential Privacy.} DP is a formal framework for quantifying privacy guarantees in randomized algorithms. It ensures that the algorithm’s output distribution changes only minimally when a single individual’s data is modified, providing robustness against arbitrary auxiliary information. We next present the standard definition of $(\varepsilon, \delta)$-differential privacy.
\begin{definition}[$(\varepsilon,\delta)$-Differential Privacy \citep{dwork2006calibrating}]
A randomized mechanism $\mathcal{M} : \mathcal{D} \rightarrow \mathcal{R}$ with domain $\mathcal{D}$ and range $\mathcal{R}$ satisfies $(\varepsilon,\delta)$-differential privacy if for any neigbouring datasets pair $D, D' \in \mathcal{D}$, and for any measurable subset of outputs $S \subseteq \mathcal{R}$, it holds that
$\mathbb{P}[\mathcal{M}(D) \in S] \;\le\; e^{\varepsilon} \cdot \mathbb{P}[\mathcal{M}(D') \in S] + \delta$
\end{definition}

\subsection{LLM reasoning for differential privacy}
A long-term goal of LLM-based reasoning for DP is to build agents that can assist throughout the full DP workflow. Ideally, given a non-private function or algorithm and a target privacy budget, such an agent would be able to design a differentially private variant or even produce a faithful implementation, with a sound argument for its privacy guarantee. Success on this broader goal would make DP more accessible to non-expert users and provide useful design baselines for DP researchers.

Achieving this end-to-end goal, however, requires a more basic capability: the ability to verify whether a proposed algorithm actually satisfies a claimed privacy guarantee. We therefore focus first on the verification task rather than on open-ended algorithm design, for two reasons. First, verification could be a necessary building block for trustworthy DP generation: a reliable DP verifier can therefore be used to evaluate candidate designs, guide iterative refinement, or serve as a reward signal for training LLMs that generate increasingly valid DP algorithms. Second, verification is practically useful in its own right. It can help audit proposed mechanisms and identify mistakes in privacy arguments.

Motivated by these considerations, in this paper we focus on the following question as the first step towards LLM-based reasoning for DP:
\textbf{Can LLMs reason about the DP guarantees of algorithms?}
Specifically, given a concrete description of a function or algorithm together with a claimed privacy guarantee, an LLM is tasked with determining whether the guarantee holds.

To support this study, we construct a benchmark with the following design principles:

\begin{itemize}[leftmargin=*, itemsep=2pt, topsep=2pt]
    \item \textbf{Broad topic coverage.} The benchmark spans core topics in DP, making the evaluation broadly representative and relevant to the DP community. A complete list of covered topics is provided in Table~\ref{tab:subject-topic-reference}.

    \item \textbf{Diverse difficulty.} It ranges from textbook mechanisms to advanced mathematical reasoning about research-level DP algorithms, enabling fine-grained assessment across difficulty levels.

    \item \textbf{Resistance to shortcut reasoning.} Instances are designed to require genuine reasoning, preventing correct answers from being obtained via recall of public training data.
\end{itemize}

Within these principles, benchmark performance provides a reliable signal of DP reasoning ability and a solid foundation for developing improved DP reasoning methods. In particular, verification is not merely a simplified proxy for algorithm generation; it is a necessary building block for any future system that aims to design, implement, and justify differentially private algorithms end to end.

\subsection{Related Work}
\label{sec:related_work}

We now discuss how the emerging direction of LLM-based reasoning for differential privacy connects to other lines of work in differential privacy. Overall, we view this emerging direction as complementary to existing directions: in some cases, they address different stages or challenges, while in others, they pursue similar goals from different angles. In both settings, combining these approaches can lead to stronger, more accessible, and reliable privacy-preserving systems.

\paragraph{DP Auditing.}
DP auditing is an important line of work for detecting implementation-level violations of differential privacy. Most existing approaches adopt a black-box framework~\citep{bichsel2021dp, ding2018detecting}: by carefully designing neighboring input datasets and empirically evaluating the outputs of a target implementation, auditors estimate a lower bound on the privacy loss. If this empirical lower bound exceeds the claimed theoretical guarantee, it indicates a likely privacy violation due to implementation bugs. Recently, several works~\citep {steinke2023privacy, mahloujifar2024auditing, xiang2025privacy} have focused on improving the efficiency of such audits, reducing the number of required executions to mitigate the substantial computational overhead, especially in settings involving large-scale deep learning models. For a comprehensive survey, we refer readers to \citep{annamalai2025hitchhiker}.

In contrast to implementation auditing, our work targets the correctness of algorithms themselves, as specified in natural language and mathematical notation. These two directions address complementary but equally essential stages of the DP pipeline. Implementation auditing serves as a final safeguard prior to deployment, whereas algorithm-level checking operates earlier in the design process, verifying whether a proposed algorithmic description satisfies differential privacy in the first place.

\paragraph{Programmatic DP verification.} 
A line of work~\citep{reed2010distance, barthe2014proving, barthe2016proving, albarghouthi2017synthesizing, zhang2017lightdp, sato2019approximate, near2021programming} investigates DP verification through program-language and formal-methods approaches, which encode algorithms in specialized languages and establish privacy guarantees using symbolic proofs such as type systems, relational Hoare logic, or coupling arguments.

Our work pursues the same goal of determining whether an algorithm satisfies a DP guarantee, but through LLM-based reasoning. While prior systems offer strong formal soundness, they require substantial expertise and are often limited in expressiveness. In contrast, LLMs can reason directly over natural-language and mathematical algorithm descriptions, lowering the barrier to use and enabling analysis of more expressive settings.

\paragraph{Verified DP implementations.}

Besides DP auditing and programmatic verification, there are also software libraries for verified DP implementations and privacy accounting, such as TensorFlow Privacy \citep{tensorflow_privacy_2019}, Opacus \citep{yousefpour2021opacus}, AutoDP \citep{autodp}, and OpenDP \citep{Shoemate_OpenDP_Library}. These tools provide human-verified building blocks for implementing DP mechanisms and support privacy accounting to improve implementation correctness. However, when new building blocks are introduced and when the algorithm design goes beyond what is representable using existing building blocks, we currently still rely on a slow and error-prone human-verification process. We believe LLMs, when used carefully, can help reduce human efforts and enhance correctness. Moreover, going beyond this work, LLMs can function as tool-using agents that make packages like OpenDP~\citep{Shoemate_OpenDP_Library} more accessible -- a natural and important future direction.

Under the broader objective of automating differential privacy, LLM-based reasoning is better suited to supporting algorithm design and first-pass expert-like checking, while verified DP libraries remain essential for deployment and serve as an additional verification safeguard.

\paragraph{LLM benchmarks on mathematical reasoning.}
Mathematical reasoning with large language models has been extensively studied in recent years, leading to the development of a wide range of evaluation benchmarks. Representative \emph{pre-college–level} benchmarks include MATH~\citep{Hendrycks}, GSM8K~\citep{cobbe2021gsm8k}, and AIME~\citep{AIME}, which primarily assess problem-solving skills in algebra, geometry, and arithmetic. Recent closed-source models with enhanced reasoning capabilities achieve near-perfect performance on these benchmarks, while open-source models continue to exhibit a noticeable performance gap. More challenging benchmarks target \emph{college-level and graduate-level} mathematics, including MathBench~\citep{liu2024mathbench}, Ghost~\citep{frieder2023mathematical}, GPQA~\citep{rein2024gpqa}, HARDMath~\citep{fan2024hardmath}, and MATH-Perturb~\citep{huang2025mathperturb}. These datasets require deeper conceptual understanding and multi-step reasoning. Even the strongest existing models still show substantial room for improvement.

Our proposed benchmark, \benchname{}, also contributes to this line of work by focusing on \emph{differential privacy reasoning}, a core topic in graduate-level applied mathematics and theoretical computer science.

\section{Dataset Construction}
In this section, we describe the construction of \benchname{}. Each question consists of \textit{a concrete algorithm description} together with \textit{a claimed privacy guarantee}, and asks \textit{whether the algorithm satisfies that guarantee}. To ensure broad coverage of privacy notions, the claimed guarantees are stated under different DP frameworks across questions, 
including approximate DP \citep{dwork2006calibrating}, Rényi DP \citep{mironov2017renyi}, zero-Concentrated DP\citep{bun2016concentrated}, and Gaussian DP\citep{dong2022gaussian}, with privacy parameters specified either as fixed constants or as functions of algorithmic parameters.

Guided by the principles of broad topic coverage and diverse difficulty levels, we organize the benchmark into two complementary categories. \catone{} focuses on foundational DP mechanisms and requires textbook-level reasoning about function sensitivity and noise calibration, which are fundamental components of nearly all DP algorithms. In contrast, \catwo{} targets advanced DP algorithms drawn from a broad range of research topics. These questions require nontrivial, algorithm-specific reasoning beyond standard textbook lookup, and therefore pose a substantially greater challenge than \catone{}. We further collect auxiliary metadata in \catwo{}, including \texttt{reference links} and \texttt{expert-written comments}, for better understanding the LLM reasoning.

\subsection{\textbf{\catone{}}: Mechanism-Level Instances with a Function Bank}
\catone{} evaluates models’ ability to reason about foundational, textbook-level DP mechanisms. We use a query-release type formulation: each mechanism is specified by a template and instantiated with a concrete query function sampled from a curated function bank.

\textbf{Mechanism.} We consider six mechanism settings: the Laplace mechanism \citep{dwork2006calibrating} under pureDP, the Gaussian mechanism under zCDP or Gaussian DP, Report-Noisy-Max with Gumbel noise (known as Exponential mechanism \citep{mcsherry2007mechanism}), Report-Noisy-Max with Laplace noise \citep{dwork2014algorithmic}, Report-Noisy-Max with Exponential noise (i.e. Permute-and-Flip mechanism \citep{mckenna2020permute, ding2021permute}). Each setting is instantiated as a mechanism template that adds calibrated noise to a query function, possibly followed by mechanism-specific transformations. As a concrete example, the Laplace mechanism is stated as below. Formal definitions of all six mechanisms are provided in the Appendix~\ref{apx:privacy_cat1}. 
\begin{theorem}[Laplace Mechanism \citep{dwork2006calibrating}] 
	Given query function $f:\mathcal{D} \rightarrow \mathbb{R}$, the laplace mechanism $\mathcal{A}(D)=f(D) + Lap(\Delta_f / \varepsilon)$ satisfies $\varepsilon$-DP, where $\mathrm{Lap}(b)$ denotes the Laplace distribution with density $p(x) = \frac{1}{2b}\exp(-|x|/b)$.
\end{theorem}

In our construction, the query function is drawn from a curated function bank.

\textbf{Function Bank.} To systematically evaluate models’ ability to reason about sensitivity calculation and noise calibration, we construct a curated function bank consisting of 49 functions, initially generated by GPT-5 and subsequently filtered by the authors, to ensure a wide range of sensitivity-reasoning difficulty. Without loss of generality, each function maps an $n$-dimensional data vector $\mathbf{x}\in[0,1]^n$ to a real-value scalar and we manually compute its \emph{tight global sensitivity} $\Delta_f$, which serves as the ground truth involving in constructing positive or negative question instances:
$$\Delta_f=\max_{\mathbf{x} \sim \mathbf{x}'\in [0, 1]^n}|f(\mathbf{x})-f(\mathbf{x}')|$$
The bank admits a broad difficulty range: from functions with immediate sensitivity, such as $f(\mathbf{x})=\sum_{i=1}^n x_i$ with $\Delta_f=1$, to functions requiring reasoning about global extrema and coordinate coupling, such as $f(\mathbf{x})=\sum_{i=1}^n |x_i-\bar{x}|$, whose tight sensitivity is $\Delta_f=\frac{2(n-1)}{n}$.

\textbf{\catone{} construction.} Given a DP mechanism and a function--sensitivity pair $(\mathcolor{blue}{g}, \mathcolor{blue}{\Delta_g})$ from the function bank, we generate a positive instance by correctly calibrating the noise to $\Delta_g$, as shown in the example below. Negative instances are constructed by under-calibrating the noise multiplier. For example, replacing $\mathrm{Lap}(\Delta_g/\varepsilon)$ with $\mathrm{Lap}(\Delta_g/(\mathcolor{red}{2}\varepsilon))$ yields a negative instance: the mechanism satisfies only $\mathcolor{red}{2}\varepsilon$-DP, rather than the claimed $\varepsilon$-DP guarantee.

\begin{tcolorbox}[
    colback=gray!5,
    colframe=cyan!75!black,
    title=\catone{}: Question Template for Laplace Mechanism,
    fonttitle=\bfseries,
    fontupper=\small
]
{Suppose $X=(x_1, \ldots, x_n)\in [0,1]^n$. Define $f(X)=\mathcolor{blue}{g}(X)$ and define $\tilde{f}(X) = f(X) + \mathrm{Lap}(\mathcolor{blue}{\Delta_g} \cdot \frac{1}{\varepsilon})$. The neighboring datasets $X$ and $X'$ differ in at most one element. Does $\tilde{f}$ satisfy $\varepsilon$-differential privacy?}
\end{tcolorbox}

To test genuine reasoning beyond theorem memorization, we instantiate questions with explicit mechanisms and concrete functions, rather than mechanism names alone. In addition, positive and negative instances are paired to differ only in a single noise-scaling term. These two approach make surface-level pattern matching insufficient for achieving high accuracy.

\subsection{\textbf{\catwo{}}: Algorithm-Level Instances from Research Literatures}
Beyond fundamental DP mechanisms, we further evaluate models on more advanced DP algorithms. Instances in this category are derived from algorithms in the research literature (Table~\ref{tab:subject-topic-reference}), together with a systematic perturbation approach to generate negative questions.

\begin{figure*}[t]
\centering

\begin{minipage}[t]{0.43\textwidth}
\vspace{0pt}
\centering
\fontsize{7}{8.5}\selectfont
\setlength{\tabcolsep}{3pt}
\begin{tabular}{lrrr}
\toprule
Topic & Yes & No & Total \\
\midrule
Privacy Accounting & 5 & 7 & 12 \\
Amplification by Subsampling & 2 & 3 & 5 \\
Private Selection & 2 & 3 & 5 \\
DP Gradient Descent & 1 & 12 & 13 \\
DP Linear Regression & 1 & 5 & 6 \\
DP Hyperparameter Tuning & 1 & 5 & 6 \\
Private Agg. of Teacher Ensembles & 1 & 2 & 3 \\
Propose-Test-Release & 2 & 2 & 4 \\
Parallel Composition & 6 & 8 & 14 \\
Sparse Vector Technique & 3 & 5 & 8 \\
Sequential/Adaptive Composition & 3 & 4 & 7 \\
Smooth Sensitivity & 2 & 1 & 3 \\
DP Adam & 1 & 2 & 3 \\
DP SGD & 3 & 6 & 9 \\
Mean Estimation & 5 & 10 & 15 \\
Objective Perturbation & 1 & 3 & 4 \\
Output Perturbation & 1 & 2 & 3 \\
Quantile Estimation & 2 & 3 & 5 \\
\midrule
Total & 42 & 83 & 125 \\
\bottomrule
\end{tabular}
\captionof{table}{Topic distribution in \catwo{}.}
\label{tab:topics_distribution}
\end{minipage}
\hfill
\begin{minipage}[t]{0.54\textwidth}
\vspace{0pt}
\centering
\fontsize{7}{8.5}\selectfont
\begin{tcolorbox}[
    colback=gray!5,
    colframe=cyan!75!black,
    fontupper=\footnotesize,
    boxsep=2pt,
    left=3pt,
    right=3pt,
    top=3pt,
    bottom=3pt
]
{\color{amber}Suppose that for all $z\in \mathcal{Z}$, $\ell(\cdot, z)$ is twice-differentiable, and the rank of its Hessian $\nabla^2\ell(\mathbf{w}, z)$ at any $\mathbf{w}\in \mathcal{W}$ is at most 1. Also assume that the smoothness parameter satisfies $\beta \leq \varepsilon n\lambda$.} 

{\color{teal}Is the following algorithm $(\varepsilon, \delta)$-differentially private?}

{
\SetAlCapFnt{\footnotesize}
\SetAlCapNameFnt{\footnotesize}
\begin{algorithm}[H]
\footnotesize
{\color{purple}
\caption{$\mathcal{A}_{\text{ObjP}}$: Objective Perturbation}
\KwIn{Private dataset $S = (z_1, \ldots, z_n) \in \mathcal{Z}^n$, 
{\color{amber}$L$-Lipschitz, $\beta$-smooth, convex loss function $\ell$, convex set 
$\mathcal{W} \subseteq \mathbb{R}^d$}, privacy parameters 
$\varepsilon \le 1$, $\delta \le \nicefrac{1}{n^2}$, regularization parameter $\lambda$.}

\BlankLine
\textbf{1:} Sample $\mathbf{G} \sim \mathcal{N}(\mathbf{0}, \sigma^2 \mathbf{I}_d)$, 
where $\sigma^2 = \dfrac{10 L^2 \log(1/\delta)}{\varepsilon^2}$\;

\textbf{2:} \textbf{return} 
$\hat{\mathbf{w}} = 
\arg\min\limits_{\mathbf{w} \in \mathcal{W}} 
\hat{\mathcal{L}}(\mathbf{w}; S) 
+ \frac{\langle \mathbf{G}, \mathbf{w} \rangle}{n} 
+ \lambda \|\mathbf{w}\|^2$, 
where $\hat{\mathcal{L}}(\mathbf{w}; S) 
\triangleq \dfrac{1}{n} \sum_{i=1}^{n} \ell(\mathbf{w}, z_i)$\;
}
\end{algorithm}
}
\end{tcolorbox}
\vspace{0.6em} 
\captionof{figure}{Example question from \catwo{} (Qid 87).}
\label{fig:cat2-positive-example}
\end{minipage}

\end{figure*}

\textbf{DP algorithm selection from the literature.}
We organize algorithm selection around four major areas in differential privacy: \emph{DP accounting}, \emph{DP statistics}, \emph{DP machine learning}, and \emph{data-adaptive mechanisms}. Within these areas, we identify 16 commonly studied topics and select representative algorithms for each. Table~\ref{tab:topics_distribution} summarizes the topics and the corresponding number of instances.%

\textbf{Structured instance construction.} For each selected algorithm, we construct both positive and negative instances. Algorithms taken from the peer-reviewed literature generally yield positive (\texttt{yes}) instances. To test whether models understand the underlying DP reasoning rather than merely memorize canonical results, we systematically perturb these algorithms to create negative (\texttt{no}) instances. Notice that each positive example is decomposed into three components: (i) {\color{purple}the algorithm description}, (ii) {\color{amber}the underlying assumptions or conditions}, and (iii) {\color{teal}the claimed DP guarantee}. We provide a colored text box for illustration (Fig.~\ref{fig:cat2-positive-example}). We then generate negative instances by perturbing exactly one of these components at a time:
(1) \emph{algorithm perturbations}, such as altering the necessary algorithmic steps;
(2) \emph{assumption perturbations}, such as removing essential conditions (e.g., convexity of the loss function); and
(3) \emph{guarantee perturbations}, such as claiming a strictly stronger privacy guarantee than what the algorithm can support. Concrete examples of these perturbations are illustrated in Appendix~\ref{appendix:example}. We also include a more fine-grained taxonomy of error patterns in the Appendix~\ref{appendix:error_taxonomy}.

\textbf{Resistance to shortcut reasoning.}
Negative instances are constructed through targeted perturbations, so models cannot succeed by simply memorizing published results. Such memorization would tend to incorrectly accept perturbed algorithms that superficially resemble valid ones. For positive instances, we conduct a rephrasing-based perturbation test. The stable accuracy suggests that model performance is not driven by memorization. We defer details to Appendix~\ref{apx:para}.

\textbf{Evaluation scope.}
This category is substantially more challenging: correctness cannot be checked by directly applying textbook DP mechanisms, but requires algorithm-specific reasoning about advanced analyses and nontrivial design choices. Since each question is associated with a topic in Table~\ref{tab:topics_distribution}, this category also supports fine-grained evaluation across DP research topics.

\textbf{Auxiliary metadata.}
Beyond questions and binary labels, we provide auxiliary metadata to support validation and future research on automated DP reasoning. Each instance includes a \texttt{reference link} pointing to the source paper underlying the instance and \texttt{expert-written comments} that serve as proof sketches. In our experiments, this metadata allows us to evaluate whether knowledge augmentation improves LLM performance, and to analyze model behavior beyond binary accuracy by checking whether the reasoning trace correctly supports the final decision. \footnote{For the better readability of metadata, please refer to: \url{https://github.com/erchiw/DPriv-Bench/blob/main/cate_2_metadata.pdf}}

\section{Experimental Setup and Results Overview}
\textbf{Rubric-guided binary evaluation.} To enable reliable and automated metric computation, we adopt a standardized prompt template that instructs the LLM to produce a binary “yes” or “no” decision in a fixed, machine-parsable format. 
If a model fails to follow this instruction and does not output an explicit binary verdict, we apply a secondary judging step using GPT-4o to map the response to a “yes” or “no” label. Each question is evaluated using the binary correctness metric. For the main results, we run five trials with different random seeds and report accuracy aggregated across seeds.

\textbf{Benchmarking LLMs.} We evaluate 11 LLMs on both \catone{} and \catwo{}. The models cover the closed-source models (\gptfive{} with minimal reasoning effort, \gptfive{} with highest reasoning effort, Gemini-3, Gemini-2.5-flash, Claude-Sonnet, Claude-Opus), the open-source models (Qwen3-30-Think, Qwen3-30-Instruct, DeepSeek-R1, DeepSeek-V3.1-chat, Goedel-Prover-V2). In addition, for \catwo{}, the advanced algorithm track, we further evaluate three up-to-date frontier models: \gemthro{}, \gptffh{}, and \gptfvh{}. We report model details, including version and release date, in Table~\ref{tab:models}. Throughout all experiments, we disabled the tool use for a fair comparison. %

In Section~\ref{sec:main}, we will present the results for the 11 LLMs on our benchmark \benchname{}. In Section~\ref{sec:analysis}, we will conduct further analysis supported by our \textit{metadata} and the case study for failure modes to better understand the model performance and gain more intuition of how to improve the performance in future work.

\section{Main Results}
\label{sec:main}

For \catone{}, we highlight two findings. First, the strongest closed-source models, \gpthigh{} and \gemhigh{}, achieve near-perfect performance, while other models leave clear room for improvement. Second, most models, including open-source ones, perform nearly perfectly on questions involving the standard Laplace mechanism, but accuracy drops substantially on the other five mechanisms. For \catwo{}, we again observe that \gpthigh{} and \gemhigh{} perform best, but both still have substantial room for improvement in this more challenging setting. We further analyze their per-topic accuracy, ranking topics by performance to identify which DP research areas are easiest and hardest for LLMs to reason about.

\vspace{-.5em}

\begin{table}[H]
\caption{Model accuracy ($\uparrow$) on the \catone{} (mean  $\pm$ $1.96\times$standard error). Model-Avg reports the averaged accuracy across the six tasks for each model. Task-Avg reports the accuracy after averaging across the eleven models for each task.}
\label{tab:cat1_overall}
\centering
\fontsize{7.5}{10}\selectfont
\begin{tabular}{lccccccc}
\toprule
 & Task-Avg & Lap. & Gauss. (GDP) & Gauss. (zCDP) & ExpMech & Lap. RNM & P-n-F \\
\midrule
\gpthigh{} & 0.995 & 1.000 $\pm$ 0.000 & 0.976 $\pm$ 0.014 & 1.000 $\pm$ 0.000 & 1.000 $\pm$ 0.000 & 0.998 $\pm$ 0.004 & 0.996 $\pm$ 0.005 \\
\gemhigh{} & 0.923 & 1.000 $\pm$ 0.000 & 0.996 $\pm$ 0.005 & 1.000 $\pm$ 0.000 & 0.959 $\pm$ 0.022 & 0.645 $\pm$ 0.019 & 0.939 $\pm$ 0.006 \\
\gemmin{} & 0.778 & 1.000 $\pm$ 0.000 & 0.743 $\pm$ 0.052 & 0.996 $\pm$ 0.005 & 0.749 $\pm$ 0.045 & 0.514 $\pm$ 0.026 & 0.665 $\pm$ 0.015 \\
\claudesonnet{} & 0.738 & 0.949 $\pm$ 0.017 & 0.814 $\pm$ 0.015 & 0.931 $\pm$ 0.022 & 0.588 $\pm$ 0.023 & 0.549 $\pm$ 0.038 & 0.600 $\pm$ 0.038 \\
\gptmin{} & 0.736 & 0.963 $\pm$ 0.022 & 0.733 $\pm$ 0.046 & 0.835 $\pm$ 0.032 & 0.782 $\pm$ 0.016 & 0.573 $\pm$ 0.019 & 0.533 $\pm$ 0.037 \\
\claudeopus{} & 0.728 & 0.955 $\pm$ 0.023 & 0.753 $\pm$ 0.038 & 0.908 $\pm$ 0.017 & 0.502 $\pm$ 0.031 & 0.590 $\pm$ 0.026 & 0.657 $\pm$ 0.035 \\
\midrule[0.1pt]
\dpskchat{} & 0.841 & 0.992 $\pm$ 0.007 & 0.951 $\pm$ 0.016 & 0.994 $\pm$ 0.005 & 0.818 $\pm$ 0.026 & 0.643 $\pm$ 0.033 & 0.651 $\pm$ 0.038 \\
\dpskrone{} & 0.741 & 0.933 $\pm$ 0.030 & 0.929 $\pm$ 0.020 & 0.943 $\pm$ 0.020 & 0.763 $\pm$ 0.043 & 0.359 $\pm$ 0.040 & 0.522 $\pm$ 0.029 \\
\qweninst{} & 0.671 & 0.992 $\pm$ 0.007 & 0.567 $\pm$ 0.038 & 0.918 $\pm$ 0.006 & 0.520 $\pm$ 0.033 & 0.508 $\pm$ 0.032 & 0.520 $\pm$ 0.054 \\
\goeprove{} & 0.628 & 0.873 $\pm$ 0.033 & 0.596 $\pm$ 0.041 & 0.822 $\pm$ 0.027 & 0.500 $\pm$ 0.035 & 0.469 $\pm$ 0.022 & 0.506 $\pm$ 0.030 \\
\qwenthink{} & 0.565 & 0.973 $\pm$ 0.015 & 0.545 $\pm$ 0.010 & 0.569 $\pm$ 0.013 & 0.418 $\pm$ 0.023 & 0.416 $\pm$ 0.037 & 0.465 $\pm$ 0.042 \\
\midrule
Model-Avg & - & 0.966 & 0.782 & 0.901 & 0.691 & 0.570 & 0.641 \\
\bottomrule
\end{tabular}
\vspace{-1em}
\end{table}

\subsection{Model Performance on \catone{} Benchmark}
\textbf{Overall performance.} Table~\ref{tab:cat1_overall} summarizes overall performance on \catone{}. Among all evaluated models, \gpthigh{} is the only one to achieve near-perfect accuracy ($0.995$). The strongest open-source model, DeepSeek-V3.1-chat, reaches $0.841$, leaving a substantial gap relative to the best closed-source models. Overall, these results show that sensitivity calculation and noise calibration remain challenging for open-source models we tested.

\textbf{Fine-grained results across mechanisms.} We further report mechanism-level accuracy in Table~\ref{tab:cat1_overall}. Nearly all models except Goedel-Prover-V2 achieve high performance ($\geq 0.94$) on the Laplace mechanism, but accuracy drops substantially on the other five mechanisms. Since all questions use the same function bank, the main difference is not sensitivity computation, but how the noise multiplier is calibrated to satisfy mechanism-specific privacy guarantees. This suggests that many models can compute sensitivity correctly yet struggle to apply it in less familiar DP mechanisms, likely because the Laplace mechanism is more prevalent in training data.

A particularly interesting case is Report-Noisy-Max with Laplace noise. \gpthigh{} is the only model with perfect accuracy, while all others, including \gemhigh{}, perform near or below the $0.5$ random-guess baseline. Inspection shows that many models incorrectly reuse the standard Laplace noise scale, adding $\mathrm{Lap}(\Delta_f/\varepsilon)$ instead of the required $\mathrm{Lap}(2\Delta_f/\varepsilon)$ to each candidate value. They therefore tend to answer ``yes'' for all instances, yielding roughly $0.5$ accuracy. This confusion between closely related mechanisms highlights a limitation in mechanism-specific DP reasoning.

\subsection{Model Performance on \catwo{} Benchmark}

\textbf{Overall performance.} In Table~\ref{tab:cat2_overall}, we report F1, precision, and recall for \catwo{}. \gptfvh{} and \gemthro{} achieve the best overall performance, with F1 scores of $0.829$ and $0.810$, respectively; \gemthro{} has higher precision, while \gptfvh{} has higher recall. Among open-source models, the DeepSeek series performs best, reaching an F1 score of $0.614$, outperforming \gptmin{} ($0.548$) and \gemmin{} ($0.602$). Nevertheless, most other models perform close to or below a naive always-yes baseline ($\mathrm{F1}\approx0.503$)\footnote{This baseline predicts ``yes'' for all \catwo{} instances, yielding $\mathrm{Recall}=1.00$, $\mathrm{Precision}=\frac{42}{42+83}\approx0.336$, and $\mathrm{F1}\approx0.503$.}, indicating that advanced algorithm-level DP reasoning remains challenging for current LLMs.

\begin{table}[t]
\centering
\fontsize{7.5}{8}\selectfont
\caption{Model performance on the \catwo{} ($\uparrow$) (mean ± 1.96×standard error).}
\label{tab:cat2_overall}
\begin{tabular}{lccc}
\toprule[1.5pt]
Model & F1 Score & Precision & Recall \\
\midrule[0.8pt]
\gptfvh{} & 0.829 $\pm$ 0.021 & 0.807 $\pm$ 0.026 & 0.852 $\pm$ 0.017 \\ 
\gemthro{}   & 0.810 $\pm$ 0.016 & 0.831 $\pm$ 0.025 & 0.790 $\pm$ 0.027 \\
\gptffh{} & 0.793 $\pm$ 0.009 & 0.731 $\pm$ 0.013 & 0.867 $\pm$ 0.019 \\
\gemhigh{} & 0.748 $\pm$ 0.019 & 0.699 $\pm$ 0.018 & 0.805 $\pm$ 0.034 \\
\gpthigh{} & 0.742 $\pm$ 0.032 & 0.716 $\pm$ 0.044 & 0.771 $\pm$ 0.024 \\
\gemmin{} & 0.602 $\pm$ 0.028 & 0.637 $\pm$ 0.038 & 0.571 $\pm$ 0.026 \\
\gptmin{} & 0.548 $\pm$ 0.036 & 0.497 $\pm$ 0.031 & 0.614 $\pm$ 0.052 \\
\claudesonnet{} & 0.518 $\pm$ 0.046 & 0.499 $\pm$ 0.040 & 0.538 $\pm$ 0.054 \\
\claudeopus{} & 0.487 $\pm$ 0.026 & 0.516 $\pm$ 0.028 & 0.462 $\pm$ 0.028 \\
\midrule[0.3pt]
\dpskrone{} & 0.614 $\pm$ 0.050 & 0.657 $\pm$ 0.045 & 0.576 $\pm$ 0.054 \\
\dpskchat{} & 0.603 $\pm$ 0.032 & 0.576 $\pm$ 0.029 & 0.633 $\pm$ 0.041 \\
\qweninst{} & 0.460 $\pm$ 0.019 & 0.526 $\pm$ 0.024 & 0.410 $\pm$ 0.027 \\
\qwenthink{} & 0.407 $\pm$ 0.046	& 0.509 $\pm$ 0.070 & 0.343 $\pm$ 0.052\\
\goeprove{} & 0.362 $\pm$ 0.045 & 0.509 $\pm$ 0.063 & 0.281 $\pm$ 0.037 \\
\bottomrule[1.5pt]
\end{tabular}
\end{table}

\textbf{Fine-grained results across topics.} We further analyze per-topic accuracy for the four best-performing models, \gptfvh{}, \gptffh{}, \gemthro{}, and \gemhigh{}. By ranking topics according to their mean accuracy across the two models, \textbf{we identify which topics are easiest and most challenging for LLMs to reason about.} As shown in Table~\ref{tab:cat2_topic}, \textit{Quantile}, \textit{DP-Adam}, \textit{Accounting}, and \textit{Private Selection} are the easiest topics, with all models achieving accuracy at least 0.95. In contrast, \textit{Smooth Sensitivity}, \textit{PATE}, \textit{Output Perturbation}, %
emerge as the most challenging topics, with average accuracies below 0.6. %

\begin{table}[t]
\centering
\fontsize{7.5}{8}\selectfont
\caption{\catwo{} topic-wise Accuracy ($\uparrow$). We report the top four and bottom three topics. (Please refer to Table~\ref{tab:cat2_complete} for complete results.)}
\label{tab:cat2_topic}
\begin{tabular}{lrrrrr}
\toprule[1.5pt]
Topic & Gemini-3-Pro & Gemini-3.1 & GPT-5.4-High & GPT-5.5-High & Avg  \\
\midrule
DP Adam & 1.00 & 1.00 & 1.00 & 1.00 & 1.00 \\
Quantile Estimation & 1.00 & 1.00 & 1.00 & 1.00 & 1.00 \\
Privacy Accounting & 0.92 & 1.00 & 1.00 & 0.98 & 0.97 \\
Private Selection & 0.80 & 1.00 & 1.00 & 1.00 & 0.95 \\
\ldots & & & \\
Output Perturbation & 0.33 & 0.87 & 0.40 & 0.67 & 0.57 \\
Private Aggregation of Teacher Ensembles (PATE) & 0.67 & 0.60 & 0.40 & 0.60 & 0.52 \\
Smooth Sensitivity & 0.33 & 0.33 & 0.33 & 0.33 & 0.33 \\
\bottomrule[1.5pt]
\end{tabular}
\end{table}

\section{More Analytic Results and Case Study}
\label{sec:analysis}

\subsection{Context-Augmented DP Reasoning}

\begin{figure}[H]
  \centering
  \includegraphics[width=0.75\linewidth]{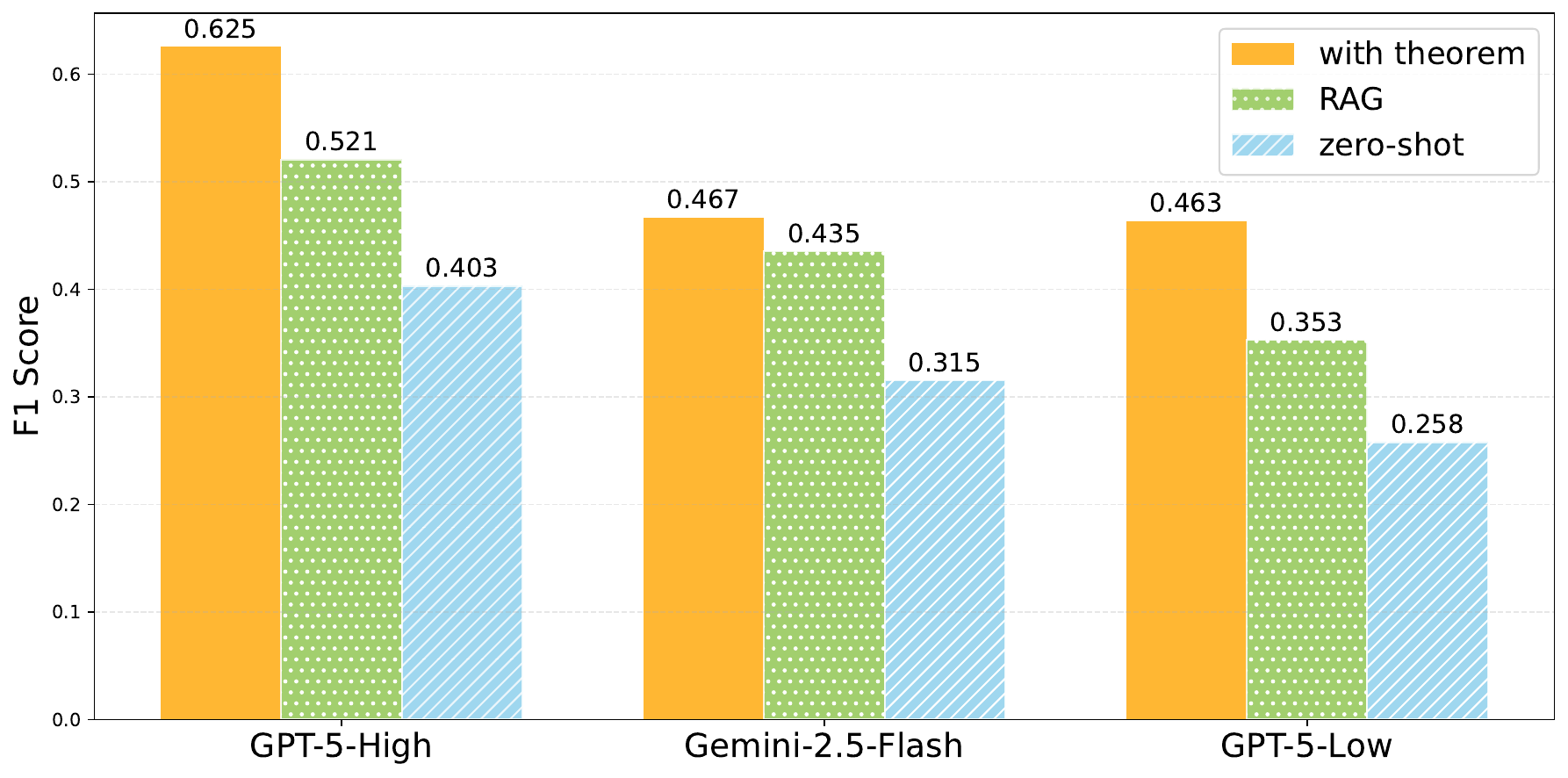}
  \caption{Performance on the 18 hardest \catwo{} questions with theorem augmentation, RAG, and zero-shot prompting. Scores are averaged over five trials.}
  \label{fig:performance_theorem}
\end{figure}

We study whether external context improves DP reasoning through theorem augmentation, retrieval augmented generation, and in-context learning. For the 18 lowest-accuracy questions in \catwo{}, 
we augment prompts with information from correct prior implementations, key theorems, and relevant definitions, either by directly injecting the information or by retrieving it from a reference-paper database via RAG. 
We augment prompts in two ways: by directly providing the relevant key theorem, or more implicitly by retrieving supporting evidence through RAG from a database of source papers identified by the metadata \texttt{reference link}.
As shown in Figure~\ref{fig:performance_theorem}, direct theorem augmentation yields the largest gains, while RAG also improves performance but more modestly, reflecting its noisier yet more realistic retrieval setting. 

We further evaluate one-shot in-context learning on the challenging Category 1 LaplaceRNM task, where instances share similar proof structures. Using a single Laplace Report Noisy Max question--answer exemplar adapted from Theorem~\ref{thm:laplaceRNM} (Appendix~\ref{apx:one-shot-temp}), we observe improved performance, as shown in Table~\ref{tab:one-shot}.

\begin{table}[H]
  \centering
  \caption{One-shot in-context learning accuracy (mean $\pm$ $1.96\times$ standard error).}
  \label{tab:one-shot}
  \begin{tabular}{lcc}
  \toprule
  Model & Zero-Shot & One-shot \\
  \midrule
  \gptmin{} & $0.573 \pm 0.019$ & $0.737 \pm 0.065$ \\
  \gemmin{} & $0.514 \pm 0.026$ & $0.661 \pm 0.073$ \\
  \bottomrule
  \end{tabular}
\end{table}

\subsection{Reasoning Plausibility and Soundness Beyond Binary Evaluation}\label{sec:response_analysis}

To assess whether LLMs reason correctly about differential privacy, rather than merely guessing the final ``yes''/``no'' label via superficial patterns or flawed reasoning, we conduct a reasoning-level evaluation on \catwo{} using a strong LLM judge, \judgemodel{}. We focus on the best-performing closed-source models: GPT and Gemini families.

Given the question, model response, and metadata \texttt{expert-written comment} as an optional reference hint, the judge assigns one of the fine-grained labels in Table~\ref{tab:softlabels}, with \emph{increasing level of correctness}: \texttt{fatal\_error}, \texttt{unclear\_or\_insufficient}, \texttt{correct\_direction\_fixable}, \texttt{minor\_gap}, \texttt{correct}. We define \emph{reasoning plausibility} as labels at least \texttt{correct\_direction\_fixable}, and \emph{reasoning soundness} as labels at least \texttt{minor\_gap}. Among responses with final predictions matches ground-truth label, we report the fraction with plausible or sound reasoning in Table~\ref{tab:truthfulness}.

\begin{wrapfigure}{r}{0.48\textwidth}
\vspace{-1.2em}

\centering
\captionsetup{font=small,skip=3pt}
\caption{Reasoning quality on \catwo{} (mean $\pm$ std).
(Note that \textit{Plausibility} and \textit{Soundness} are computed only among responses with correct final answers. See Appendix~\ref{sec:reasoningmetric} for detailed metric definitions.)
}
\label{tab:truthfulness}
\fontsize{7.5}{9}\selectfont
\begin{tabular}{llll}
\toprule
Model & Accuracy($\uparrow$) & Plausibility($\uparrow$) & Soundness($\uparrow$) \\
\midrule
\gptfvh{} & \textbf{0.882} ± 0.017 & \textbf{0.998} ± 0.004 & \textbf{0.978} ± 0.008 \\
\gptffh{} & 0.848 ± 0.008 & 0.996 ± 0.005 & 0.949 ± 0.018 \\
\gemthro{} & 0.875 ± 0.012 & 0.973 ± 0.006 & 0.848 ± 0.018 \\
\gpthigh{} & 0.819 ± 0.029 & 0.950 ± 0.019 & 0.773 ± 0.013 \\
\gemhigh{} & 0.818 ± 0.014 & 0.943 ± 0.017 & 0.732 ± 0.020 \\
\gptmin{} & 0.661 ± 0.030 & 0.874 ± 0.007 & 0.623 ± 0.022 \\
\gemmin{} & 0.746 ± 0.024 & 0.702 ± 0.037 & 0.429 ± 0.014 \\
\bottomrule
\end{tabular}

\end{wrapfigure}

We find that predictions from strong reasoning models are largely trustworthy at the level of high-level DP principles. For \gptfvh{} and \gemthro{}, both plausibility and soundness scores are close to 1. For both \gpthigh{} and \gemhigh{}, plausibility is close to one. However, their lower soundness scores show that many proofs remain incomplete or mathematically unsound. This gap is larger for weaker models; for example, \gemmin{} achieves relatively high final-answer accuracy but much lower plausibility and soundness. Overall, final-answer correctness is informative for strong reasoning models, but can overestimate proof-level reasoning quality, especially for weaker models. For examples of judge results, please refer to Appendix~\ref{apx:llm-as-judge}.

\vspace{1em}
\subsection{Case Study for Failure Modes}

\begin{figure}[t]
\centering

\begin{tcolorbox}[
    width=\linewidth,
    colback=gray!5,
    colframe=cyan!75!black,
    title=\catwo{} Question 106 (Label: No),
    fonttitle=\bfseries,
    fontupper=\small,
    boxsep=1pt,
    left=3pt,
    right=3pt,
    top=3pt,
    bottom=3pt
]
Consider the following procedure:
$\mathcal{M} := \{\mathcal{M}_i(X_i)\}_{i=1}^n$,
where $X_i \cap X_{i+1} = \emptyset$ for all
$i\in\{1,\ldots,n-1\}$, and their union covers the entire dataset,
i.e., $\bigcup_{i=1}^n X_i = X$.
Furthermore, the partition is data-independent.

Suppose each mechanism $\mathcal{M}_i$ satisfies
$(\varepsilon_i, \delta_i)$-differential privacy.
Does the combined mechanism $\mathcal{M}$ satisfy
$(\max_{i \in [n]}\varepsilon_i,\max_{i \in [n]} \delta_i)$-differential privacy?
\end{tcolorbox}

\vspace{0.5em}

\begin{tcolorbox}[
    width=\linewidth,
    colback=gray!5,
    colframe=cyan!75!black,
    title=\catwo{} Question 76 (Label: No),
    fonttitle=\bfseries,
    fontupper=\small,
    boxsep=1pt,
    left=3pt,
    right=3pt,
    top=3pt,
    bottom=3pt
]
Does the following algorithm satisfy $\varepsilon$-differential privacy
under the add/remove neighboring relation?

\vspace{2pt}
\hrule
\vspace{2pt}
\textbf{Algorithm 2: Mean estimation}
\vspace{2pt}
\hrule
\vspace{2pt}

\textbf{Input:} scalar dataset $D:=\{x_1,\ldots,x_n\}$,
privacy budget $\varepsilon$, clipping threshold $C$.

Sample $\xi_1 \sim \mathrm{Lap}(2C/\varepsilon)$.

Sample $\xi_2 \sim \mathrm{Lap}(2/\varepsilon)$.

Let
\[
\hat \theta \leftarrow
\frac{
\sum_{i=1}^{n-1} \min\{\max\{-C, x_i\},C\} + \xi_1
}{
n + \xi_2
}.
\]

\textbf{return} $\hat \theta$.
\end{tcolorbox}

\captionsetup{font=small}
\caption{Examples where LLMs confuse superficially similar privacy reasoning patterns.}
\label{fig:similar-structure-failures}

\end{figure}

We analyze two common failure modes exhibited by the top performing models. By identifying these error patterns, we aim to provide insights that can inform future method development -- for example, by using these failure modes as highlighted aspects when assessing the correctness of a reasoning trajectory. 

\textbf{Failure to identify subtle but semantically significant changes.} 
We find that LLMs can confuse questions with similar structures, leading to systematic errors. In Q106 from our dataset, we modify the standard assumption required for parallel composition: instead of requiring pairwise disjointness, i.e., $X_i \cap X_j = \emptyset$ for all $i \neq j$, we require only sequential disjointness, i.e., $X_i \cap X_{i+1} = \emptyset$. We then ask whether the resulting algorithm satisfies parallel composition. Among the models evaluated, only \gemthro{} succeeds in one out of five trials; all other models, including \gptffh{}, \gptfvh{}, \gemhigh{}, and \gpthigh{}, achieve zero accuracy on this question.

We further include Q76. In this case, the issue is that the summation in the numerator ranges from $1$ to $n-1$ rather than from $1$ to $n$, which doubles the sensitivity. As a result, the added Laplace noise is insufficient. Only \gemhigh{} and \gpthigh{} succeed in one out of five trials, while all other models, including \gptffh{}, \gptfvh{}, and \gemthro{}, achieve zero accuracy.

\textbf{Hallucination of assumptions learned from training data.} LLMs may also hallucinate assumptions based on their
pretraining data. For example, in one output perturbation question (Figure~\ref{fig:output}), the original result appears as Proposition~2.1 in \citep{lowy2021output}, where the sensitivity of the objective is defined with respect to the minimizer. In our question, however, we omit this assumption and state only that an $L_2$ sensitivity is considered in general, which should typically be interpreted as the sensitivity of the function $f$. This interpretation is insufficient to guarantee the correctness of the output perturbation algorithm. Nevertheless, the LLMs often verify the statement as correct, implicitly assuming that the sensitivity is taken with respect to the minimizer, as evidenced by their outputs. Over five trials on this question, Gemini-3-Pro consistently predicts the incorrect affirmative answer, achieving 0\% accuracy, which reflects a systematic substitution of objective sensitivity for minimizer sensitivity. GPT-5.4-High performs only slightly better, with 20\% accuracy, while GPT-5.5-High reaches 40\%. Gemini-3.1 and GPT-5-High achieve the best observed performance, but still only reach 60\% accuracy.

\begin{figure}[t]
    \begin{tcolorbox}[
        colback=gray!5,
        colframe=cyan!75!black,
        title=\catwo{} Question 91 (Label: No),
        fonttitle=\bfseries,
        fontupper=\footnotesize,
        boxsep=1pt,
        left=2pt,
        right=2pt,
        top=2pt,
        bottom=2pt
    ]

Let $\varepsilon > 0$, $\delta \in [0, \frac{1}{2})$. Let $F : \mathbb{R}^d \times \mathcal{X}^n \to \mathbb{R}$ be a function with $L_2$ sensitivity $\Delta_F$ and unique minimizer $w^*(X) = \arg\min_{w \in \mathbb{R}^d} F(w, X)$ for all $X \in \mathcal{X}^n$. The (conceptual) output perturbation algorithms $\mathcal{A}(X) = w_{\mathcal{A}}(X) := w^*(X) + z$ with noise given by density

\[
p_z(t) \propto
\begin{cases}
\exp\!\left(-\dfrac{\varepsilon \|t\|_2}{\Delta_F}\right),
& \text{if } \delta = 0, \\[4pt]
\exp\!\left(
-\dfrac{\varepsilon^2 \|t\|_2^2}
{\Delta_F^2 \left(c_\delta + \sqrt{c_\delta^2 + \varepsilon}\right)^2}
\right),
& \text{if } \delta \in \left(0, \tfrac{1}{2}\right).
\end{cases}
\]
where
\[
c_\delta := \sqrt{
\log\!\left(
\dfrac{2}{\sqrt{16\delta^2 + 1} - 1}
\right)
}.
\]
Is $\mathcal{A}(X)$ $(\varepsilon, \delta)$-differentially private?
    \end{tcolorbox}
\caption{Example Question where LLM hallucinates assumptions}
\label{fig:output}
\end{figure}

\section{Conclusion and Future Work}

\textbf{Conclusion.} We introduce \benchname{}, the first benchmark for evaluating LLM reasoning about differential privacy across both foundational and advanced DP algorithms. Our results show that leading models handle textbook DP mechanisms well but struggle with algorithm-specific analyses requiring careful accounting and assumption validation, revealing a clear gap to expert-level DP reasoning. 

\textbf{Future work.} Our work motivates improving LLM-based DP reasoning using \benchname{} as a guide. Promising next steps include \textit{automated DP algorithm generation}, which synthesizes private algorithms from non-private specifications, and \textit{implementation-level automation}, which detects and prevents privacy violations in code.

\section{Acknowledgment}

This work was supported in part by the ONR under grants N000142412304 and N00014-25-1-2116,
by the NSF under grants CNS 2048091, CIF-2402817 and CNS-2241100, and by the ARO-MURI under grant
W911NF2110317.  We acknowledge an OpenAI security research grant for providing the necessary credits and API access to their models.

\newpage

\bibliography{example_paper}

@inproceedings{dwork2006calibrating,
  title={Calibrating noise to sensitivity in private data analysis},
  author={Dwork, Cynthia and McSherry, Frank and Nissim, Kobbi and Smith, Adam},
  booktitle={Theory of cryptography conference},
  pages={265--284},
  year={2006},
  organization={Springer}
}

@inproceedings{mironov2017renyi,
  title={R{\'e}nyi differential privacy},
  author={Mironov, Ilya},
  booktitle={2017 IEEE 30th computer security foundations symposium (CSF)},
  pages={263--275},
  year={2017},
  organization={IEEE}
}

@article{dong2022gaussian,
  title={Gaussian differential privacy},
  author={Dong, Jinshuo and Roth, Aaron and Su, Weijie J},
  journal={Journal of the Royal Statistical Society Series B: Statistical Methodology},
  volume={84},
  number={1},
  pages={3--37},
  year={2022},
  publisher={Oxford University Press}
}

@inproceedings{bun2016concentrated,
  title={Concentrated differential privacy: Simplifications, extensions, and lower bounds},
  author={Bun, Mark and Steinke, Thomas},
  booktitle={Theory of cryptography conference},
  pages={635--658},
  year={2016},
  organization={Springer}
}

@article{ding2021permute,
  title={The permute-and-flip mechanism is identical to report-noisy-max with exponential noise},
  author={Ding, Zeyu and Kifer, Daniel and Steinke, Thomas and Wang, Yuxin and Xiao, Yingtai and Zhang, Danfeng and others},
  journal={arXiv preprint arXiv:2105.07260},
  year={2021}
}

@article{mckenna2020permute,
  title={Permute-and-flip: A new mechanism for differentially private selection},
  author={McKenna, Ryan and Sheldon, Daniel R},
  journal={Advances in Neural Information Processing Systems},
  volume={33},
  pages={193--203},
  year={2020}
}

@inproceedings{dong2020optimal,
  title={Optimal Differential Privacy Composition for Exponential Mechanisms},
  author={Dong, Jinshuo and Durfee, David and Rogers, Ryan},
  journal={Proceedings of the 37th International Conference on Machine Learning},
  year={2020}
}

@inproceedings{durfee2019practical,
  title={Practical Differentially Private Top-k Selection with Pay-what-you-get Composition},
  author={Durfee, David and Rogers, Ryan},
  journal={Advances in Neural Information Processing Systems},
  year={2019}
}

@inproceedings{mcsherry2007mechanism,
  title={Mechanism design via differential privacy},
  author={McSherry, Frank and Talwar, Kunal},
  booktitle={48th Annual IEEE Symposium on Foundations of Computer Science (FOCS'07)},
  pages={94--103},
  year={2007},
  organization={IEEE}
}

@article{lowy2021output,
  title={Output perturbation for differentially private convex optimization: Faster and more general},
  author={Lowy, Andrew and Razaviyayn, Meisam},
  journal={arXiv preprint arXiv:2102.04704},
  year={2021}
}

@article{dwork2014algorithmic,
  title={The algorithmic foundations of differential privacy},
  author={Dwork, Cynthia and Roth, Aaron and others},
  journal={Foundations and trends{\textregistered} in theoretical computer science},
  volume={9},
  number={3--4},
  pages={211--407},
  year={2014},
  publisher={Now Publishers, Inc.}
}

@inproceedings{abowd2018us,
  title={The US Census Bureau adopts differential privacy},
  author={Abowd, John M},
  booktitle={Proceedings of the 24th ACM SIGKDD international conference on knowledge discovery \& data mining},
  pages={2867--2867},
  year={2018}
}

@article{annamalai2025hitchhiker,
  title={The Hitchhiker's Guide to Efficient, End-to-End, and Tight DP Auditing},
  author={Annamalai, Meenatchi Sundaram Muthu Selva and Balle, Borja and Hayes, Jamie and Kaissis, Georgios and De Cristofaro, Emiliano},
  journal={arXiv preprint arXiv:2506.16666},
  year={2025}
}

@misc{google_dp_blog_2021,
  title        = {How we're helping developers with differential privacy},
  author       = {{Google Developers Blog}},
  year         = {2021},
  month        = {Jan},
  day          = {28},
  url          = {https://developers.googleblog.com/how-were-helping-developers-with-differential-privacy/},
}

@misc{nist2020,
  title        = {How we're helping developers with differential privacy},
  author       = {Simson Garfinkel},
  year         = {2020},
  month        = {Jan},
  day          = {27},
  url          = {https://csrc.nist.gov/presentations/2020/stppa1-census},
}

@misc{meta_pets_adtech_2025,
  title={How Meta Uses Privacy-Enhancing Technologies in Advertising and Analytics},
  author={Figas, Natalia},
  year={2025},
  url={https://www.avenga.com/magazine/how-meta-uses-privacy-enhancing-technologies-pets-in-adtech/},
}

@misc{icml2021_virtual_11631,
  title        = {Deploying Differential Privacy in Industry: Progress and Learnings},
  author       = {Ryan Rogers},
  year         = {2021},
  url          = {https://icml.cc/virtual/2021/11631},
}

@misc{apple_dp_aggregate_trends,
  title        = {Understanding Aggregate Trends for Apple Intelligence Using Differential Privacy},
  author       = {{Apple Blog}},
  year         = {2025},
  url          = {https://machinelearning.apple.com/research/differential-privacy-aggregate-trends},
}

@inproceedings{reed2010distance,
  title={Distance makes the types grow stronger: a calculus for differential privacy},
  author={Reed, Jason and Pierce, Benjamin C},
  booktitle={Proceedings of the 15th ACM SIGPLAN international conference on Functional programming},
  pages={157--168},
  year={2010}
}

@inproceedings{barthe2014proving,
  title={Proving differential privacy in Hoare logic},
  author={Barthe, Gilles and Gaboardi, Marco and Arias, Emilio Jes{\'u}s Gallego and Hsu, Justin and Kunz, C{\'e}sar and Strub, Pierre-Yves},
  booktitle={2014 IEEE 27th Computer Security Foundations Symposium},
  pages={411--424},
  year={2014},
  organization={IEEE}
}

@inproceedings{barthe2016proving,
  title={Proving differential privacy via probabilistic couplings},
  author={Barthe, Gilles and Gaboardi, Marco and Gr{\'e}goire, Benjamin and Hsu, Justin and Strub, Pierre-Yves},
  booktitle={Proceedings of the 31st Annual ACM/IEEE Symposium on Logic in Computer Science},
  pages={749--758},
  year={2016}
}

@article{albarghouthi2017synthesizing,
  title={Synthesizing coupling proofs of differential privacy},
  author={Albarghouthi, Aws and Hsu, Justin},
  journal={Proceedings of the ACM on Programming Languages},
  volume={2},
  number={POPL},
  pages={1--30},
  year={2017},
  publisher={ACM New York, NY, USA}
}

@inproceedings{sato2019approximate,
  title={Approximate span liftings: Compositional semantics for relaxations of differential privacy},
  author={Sato, Tetsuya and Barthe, Gilles and Gaboardi, Marco and Hsu, Justin and Katsumata, Shin-ya},
  booktitle={2019 34th Annual ACM/IEEE Symposium on Logic in Computer Science (LICS)},
  pages={1--14},
  year={2019},
  organization={IEEE}
}

@article{mahloujifar2024auditing,
  title={Auditing $ f $-Differential Privacy in One Run},
  author={Mahloujifar, Saeed and Melis, Luca and Chaudhuri, Kamalika},
  journal={arXiv preprint arXiv:2410.22235},
  year={2024}
}

@article{steinke2023privacy,
  title={Privacy auditing with one (1) training run},
  author={Steinke, Thomas and Nasr, Milad and Jagielski, Matthew},
  journal={Advances in Neural Information Processing Systems},
  volume={36},
  pages={49268--49280},
  year={2023}
}

@article{huang2025winning,
  title={Winning gold at imo 2025 with a model-agnostic verification-and-refinement pipeline},
  author={Huang, Yichen and Yang, Lin F},
  journal={arXiv preprint arXiv:2507.15855},
  year={2025}
}

@misc{openai_gpt5_announcement,
  title        = {Introducing {GPT-5}},
  author       = {{OpenAI}},
  year         = {2025},
  url          = {https://openai.com/index/introducing-gpt-5/},
}

@misc{deepmind_gemini3pro_modelcard,
  title        = {Gemini 3 Pro Model Card},
  author       = {{Google DeepMind}},
  year         = {2025},
  url          = {https://storage.googleapis.com/deepmind-media/Model-Cards/Gemini-3-Pro-Model-Card.pdf},
}

@inproceedings{Hendrycks,
 author = {Hendrycks, Dan and Burns, Collin and Kadavath, Saurav and Arora, Akul and Basart, Steven and Tang, Eric and Song, Dawn and Steinhardt, Jacob},
 booktitle = {Proceedings of the Neural Information Processing Systems Track on Datasets and Benchmarks},
 editor = {J. Vanschoren and S. Yeung},
 pages = {},
 title = {Measuring Mathematical Problem Solving With the MATH Dataset},
 volume = {1},
 year = {2021}
}

@article{lyu2017understanding,
  title={Understanding the sparse vector technique for differential privacy},
  author={Lyu, Min and Su, Dong and Li, Ninghui},
  journal={Proceedings of the VLDB Endowment},
  volume={10},
  number={6},
  pages={637--648},
  year={2017},
  publisher={VLDB Endowment}
}

@article{chu2025dpcheatsheet,
  title={DPCheatSheet: Using Worked and Erroneous LLM-usage Examples to Scaffold Differential Privacy Implementation},
  author={Chu, Shao-Yu and Tian, Yuhe and Wang, Yu-Xiang and Jin, Haojian},
  journal={arXiv preprint arXiv:2509.12590},
  year={2025}
}

@article{cobbe2021gsm8k,
  title={Training Verifiers to Solve Math Word Problems},
  author={Cobbe, Karl and Kosaraju, Vineet and Bavarian, Mohammad and Chen, Mark and Jun, Heewoo and Kaiser, Lukasz and Plappert, Matthias and Tworek, Jerry and Hilton, Jacob and Nakano, Reiichiro and Hesse, Christopher and Schulman, John},
  journal={arXiv preprint arXiv:2110.14168},
  year={2021}
}

@misc{AIME,
  title        = {American Invitational Mathematics Examination},
  author       = {{Mathematical Association of America}},
  year         = {2025},
  howpublished = {\url{https://maa.org/maa-invitational-competitions/}},
  note         = {Accessed: 2026-01-23}
}

@inproceedings{
huang2025mathperturb,
title={{MATH}-Perturb: Benchmarking {LLM}s' Math Reasoning Abilities against Hard Perturbations},
author={Kaixuan Huang and Jiacheng Guo and Zihao Li and Xiang Ji and Jiawei Ge and Wenzhe Li and Yingqing Guo and Tianle Cai and Hui Yuan and Runzhe Wang and Yue Wu and Ming Yin and Shange Tang and Yangsibo Huang and Chi Jin and Xinyun Chen and Chiyuan Zhang and Mengdi Wang},
booktitle={Forty-second International Conference on Machine Learning},
year={2025},
url={https://openreview.net/forum?id=OZy70UggXr}
}

@inproceedings{
frieder2023mathematical,
title={Mathematical Capabilities of Chat{GPT}},
author={Simon Frieder and Luca Pinchetti and Alexis Chevalier and Ryan-Rhys Griffiths and Tommaso Salvatori and Thomas Lukasiewicz and Philipp Christian Petersen and Julius Berner},
booktitle={Thirty-seventh Conference on Neural Information Processing Systems Datasets and Benchmarks Track},
year={2023},
url={https://openreview.net/forum?id=xJ7YWXQOrg}
}

@inproceedings{
rein2024gpqa,
title={{GPQA}: A Graduate-Level Google-Proof Q\&A Benchmark},
author={David Rein and Betty Li Hou and Asa Cooper Stickland and Jackson Petty and Richard Yuanzhe Pang and Julien Dirani and Julian Michael and Samuel R. Bowman},
booktitle={First Conference on Language Modeling},
year={2024},
url={https://openreview.net/forum?id=Ti67584b98}
}

@article{fan2024hardmath,
  title={Hardmath: A benchmark dataset for challenging problems in applied mathematics},
  author={Fan, Jingxuan and Martinson, Sarah and Wang, Erik Y and Hausknecht, Kaylie and Brenner, Jonah and Liu, Danxian and Peng, Nianli and Wang, Corey and Brenner, Michael P},
  journal={arXiv preprint arXiv:2410.09988},
  year={2024}
}

@article{liu2024mathbench,
  title={Mathbench: Evaluating the theory and application proficiency of llms with a hierarchical mathematics benchmark},
  author={Liu, Hongwei and Zheng, Zilong and Qiao, Yuxuan and Duan, Haodong and Fei, Zhiwei and Zhou, Fengzhe and Zhang, Wenwei and Zhang, Songyang and Lin, Dahua and Chen, Kai},
  journal={arXiv preprint arXiv:2405.12209},
  year={2024}
}

@inproceedings{bichsel2021dp,
  title={Dp-sniper: Black-box discovery of differential privacy violations using classifiers},
  author={Bichsel, Benjamin and Steffen, Samuel and Bogunovic, Ilija and Vechev, Martin},
  booktitle={2021 IEEE Symposium on Security and Privacy (SP)},
  pages={391--409},
  year={2021},
  organization={IEEE}
}

@inproceedings{ding2018detecting,
  title={Detecting violations of differential privacy},
  author={Ding, Zeyu and Wang, Yuxin and Wang, Guanhong and Zhang, Danfeng and Kifer, Daniel},
  booktitle={Proceedings of the 2018 ACM SIGSAC Conference on Computer and Communications Security},
  pages={475--489},
  year={2018}
}

@inproceedings{xiang2025privacy,
  title={Privacy Audit as Bits Transmission:(Im) possibilities for Audit by One Run},
  author={Xiang, Zihang and Wang, Tianhao and Wang, Di},
  booktitle={USENIX Security},
  year={2025}
}

@inproceedings{zhang2017lightdp,
  title={LightDP: Towards automating differential privacy proofs},
  author={Zhang, Danfeng and Kifer, Daniel},
  booktitle={Proceedings of the 44th ACM SIGPLAN Symposium on Principles of Programming Languages},
  pages={888--901},
  year={2017}
}

@article{near2021programming,
  title={Programming differential privacy},
  author={Near, Joseph P and Abuah, Chik{\'e}},
  journal={URL: https://uvm},
  year={2021}
}

@article{chaudhuri2011differentially,
  title={Differentially private empirical risk minimization.},
  author={Chaudhuri, Kamalika and Monteleoni, Claire and Sarwate, Anand D},
  journal={Journal of Machine Learning Research},
  volume={12},
  number={3},
  year={2011}
}

@article{arxiv14057085,
  title={Differentially Private Empirical Risk Minimization: Efficient Algorithms and Tight Error Bounds},
  author={Raef Bassily and Adam Smith and Abhradeep Thakurta},
  journal={arXiv preprint arXiv:1405.7085},
  year={2014}
}

@article{arxiv170207476,
  title={Renyi Differential Privacy},
  author={Ilya Mironov},
  journal={arXiv preprint arXiv:1702.07476},
  year={2017}
}

@article{arxiv180208908,
  title={Scalable Private Learning with PATE},
  author={Nicolas Papernot and Shuang Song and Ilya Mironov and Ananth Raghunathan and Kunal Talwar and Úlfar Erlingsson},
  journal={arXiv preprint arXiv:1802.08908},
  year={2018}
}

@article{arxiv180701647,
  title={Privacy Amplification by Subsampling: Tight Analyses via Couplings and Divergences},
  author={Borja Balle and Gilles Barthe and Marco Gaboardi},
  journal={arXiv preprint arXiv:1807.01647},
  year={2018}
}

@article{arxiv190602830,
  title={Average-Case Averages: Private Algorithms for Smooth Sensitivity and Mean Estimation},
  author={Mark Bun and Thomas Steinke},
  journal={arXiv preprint arXiv:1906.02830},
  year={2019}
}

@article{arxiv190809970,
  title={Private Stochastic Convex Optimization with Optimal Rates},
  author={Raef Bassily and Vitaly Feldman and Kunal Talwar and Abhradeep Thakurta},
  journal={arXiv preprint arXiv:1908.09970},
  year={2019}
}

@article{arxiv190913830,
  title={Optimal Differential Privacy Composition for Exponential Mechanisms and the Cost of Adaptivity},
  author={Jinshuo Dong and David Durfee and Ryan Rogers},
  journal={arXiv preprint arXiv:1909.13830},
  year={2019}
}

@article{arxiv200400010,
  title={The Discrete Gaussian for Differential Privacy},
  author={Clément L. Canonne and Gautam Kamath and Thomas Steinke},
  journal={arXiv preprint arXiv:2004.00010},
  year={2020}
}

@article{arxiv200407223,
  title={Bounding, Concentrating, and Truncating: Unifying Privacy Loss Composition for Data Analytics},
  author={Mark Cesar and Ryan Rogers},
  journal={arXiv preprint arXiv:2004.07223},
  year={2020}
}

@article{arxiv211003620,
  title={Hyperparameter Tuning with Renyi Differential Privacy},
  author={Nicolas Papernot and Thomas Steinke},
  journal={arXiv preprint arXiv:2110.03620},
  year={2021}
}

@article{arxiv211005679,
  title={Large Language Models Can Be Strong Differentially Private Learners},
  author={Xuechen Li and Florian Tramèr and Percy Liang and Tatsunori Hashimoto},
  journal={arXiv preprint arXiv:2110.05679},
  year={2021}
}

@article{arxiv220413650,
  title={Unlocking High-Accuracy Differentially Private Image Classification through Scale},
  author={Soham De and Leonard Berrada and Jamie Hayes and Samuel L. Smith and Borja Balle},
  journal={arXiv preprint arXiv:2204.13650},
  year={2022}
}

@article{arxiv230100301,
  title={Generalized PTR: User-Friendly Recipes for Data-Adaptive Algorithms with Differential Privacy},
  author={Rachel Redberg and Yuqing Zhu and Yu-Xiang Wang},
  journal={arXiv preprint arXiv:2301.00301},
  year={2022}
}

@article{arxiv231206658,
  title={Mean estimation in the add-remove model of differential privacy},
  author={Alex Kulesza and Ananda Theertha Suresh and Yuyan Wang},
  journal={arXiv preprint arXiv:2312.06658},
  year={2023}
}

@article{rehn2025optimal,
  title={On Optimal Hyperparameters for Differentially Private Deep Transfer Learning},
  author={Rehn, Aki and Zhao, Linzh and Heikkil{\"a}, Mikko A and Honkela, Antti},
  journal={arXiv preprint arXiv:2510.20616},
  year={2025}
}

@article{harrison2025exact,
  title={Exact zCDP Characterizations for Fundamental Differentially Private Mechanisms},
  author={Harrison, Charlie and Manurangsi, Pasin},
  journal={arXiv preprint arXiv:2510.25746},
  year={2025}
}

@article{McSherry_2010,
  title={Privacy integrated queries: an extensible platform for privacy-preserving data analysis},
  author={McSherry, Frank},
  journal={Communications of the ACM},
  year={2010}
}

@inproceedings{Abadi_2016,
  title={Deep Learning with Differential Privacy},
  author={Abadi, Martin and Chu, Andy and Goodfellow, Ian and McMahan, H. Brendan and Mironov, Ilya and Talwar, Kunal and Zhang, Li},
  journal={Proceedings of the 2016 ACM SIGSAC Conference on Computer and Communications Security},
  year={2016}
}

@InProceedings{pmlr-v162-kaplan22a,
  title={Differentially Private Approximate Quantiles},
  author={Kaplan, Haim and Schnapp, Shachar and Stemmer, Uri},
  journal={Proceedings of the 39th International Conference on Machine Learning},
  year={2022}
}

@InProceedings{pmlr-v23-kifer12,
  title={Private Convex Empirical Risk Minimization and High-dimensional Regression},
  author={Kifer, Daniel and Smith, Adam and Thakurta, Abhradeep},
  journal={Proceedings of the 25th Annual Conference on Learning Theory},
  year={2012}
}

@InProceedings{pmlr-v70-sheffet17a,
  title={Differentially Private Ordinary Least Squares},
  author={Or Sheffet},
  journal={Proceedings of the 34th International Conference on Machine Learning},
  year={2017}
}

@inproceedings{dwork2009differential,
  title={Differential Privacy and Robust Statistics},
  author={Dwork, Cynthia and Lei, Jing},
  journal={Proceedings of the 41st Annual ACM Symposium on Theory of Computing (STOC)},
  year={2009}
}

@article{liu2022identification,
  title={Identification, amplification and measurement: A bridge to gaussian differential privacy},
  author={Liu, Yi and Sun, Ke and Jiang, Bei and Kong, Linglong},
  journal={Advances in Neural Information Processing Systems},
  volume={35},
  pages={11410--11422},
  year={2022}
}

@incollection{vadhan2017complexity,
  title={The complexity of differential privacy},
  author={Vadhan, Salil},
  booktitle={Tutorials on the Foundations of Cryptography: Dedicated to Oded Goldreich},
  pages={347--450},
  year={2017},
  publisher={Springer}
}

@software{Shoemate_OpenDP_Library,
author = {Shoemate, Michael and Vyrros, Andrew and McCallum, Chuck and Prasad, Raman and Durbin, Philip and Casacuberta Puig, Sílvia and Cowan, Ethan and Xu, Vicki and Ratliff, Zachary and Berrios, Nicolás and Whitworth, Alex and Eliot, Michael and Lebeda, Christian and Renard, Oren and McKay Bowen, Claire},
license = {MIT},
title = {{OpenDP Library}},
url = {https://github.com/opendp/opendp}
}

@misc{autodp,
  author = {{Autodp Contributors.}},
  title        = {autodp: Automating differential privacy computation},
  year         = {2023},
  howpublished = {\url{https://github.com/yuxiangw/autodp}},
  note         = {GitHub repository, version 0.2.3.1, accessed 2026-04-16}
}

@article{yousefpour2021opacus,
  title={Opacus: User-friendly differential privacy library in PyTorch},
  author={Yousefpour, Ashkan and Shilov, Igor and Sablayrolles, Alexandre and Testuggine, Davide and Prasad, Karthik and Malek, Mani and Nguyen, John and Ghosh, Sayan and Bharadwaj, Akash and Zhao, Jessica and others},
  journal={arXiv preprint arXiv:2109.12298},
  year={2021}
}

@misc{tensorflow_privacy_2019,
  author       = {{Tensorflow Privacy Contributors}},
  title        = {{TensorFlow Privacy: Library for training machine learning models with privacy for training data}},
  year         = {2019},
  howpublished = {\url{https://github.com/tensorflow/privacy}},
  note         = {GitHub repository, accessed 2026-04-16}
}
\bibliographystyle{plain}

\newpage
\appendix
\tableofcontents
\onecolumn
\section{Privacy Guarantee and References}
\subsection{Privacy guarantee for \catone{}}\label{apx:privacy_cat1}
In this section, we state formal proof for all six mechanisms use in \catone{}. Through the question statement in \catone{}, we use \emph{replace-one} neighbouring relationship.
\begin{theorem}[Laplace Mechanism \citep{dwork2006calibrating}] 
Let $f : \mathcal X^n \to \mathbb R^d$ be a function with $\ell_1$-sensitivity $\Delta_1(f) := \max\limits_{X \sim X^\prime} \| f(X) - f(X') \|_1$.
The Laplace mechanism defined as follow satisfies $\varepsilon$-DP: 
\[
\mathcal M(X) = f(X) + Z,
\qquad
Z_i \stackrel{\text{i.i.d.}}{\sim} \mathrm{Lap}\!\left(\frac{\Delta_1(f)}{\varepsilon}\right).
\]
\end{theorem}

\begin{theorem}[zCDP guarantee of Gaussian Mechanism \citep{bun2016concentrated}] 
Let $f : \mathcal X^n \to \mathbb R^d$ be a function with $\ell_2$-sensitivity $\Delta_2(f) := \max\limits_{X \sim X^\prime} \| f(X) - f(X') \|_2$.
The Gaussian mechanism defined as follow satisfies $\rho$-zero Concentrated Differential Privacy (zCDP): 
\[
\mathcal M(X) = f(X) + Z,
\qquad
Z_i \stackrel{\text{i.i.d.}}{\sim} \mathcal{N}\!\left(0, \frac{\Delta_2(f)^2}{2\rho}\right).
\]
\end{theorem}

\begin{theorem}[GDP guarantee of Gaussian Mechanism \citep{dong2022gaussian}] 
Let $f : \mathcal X^n \to \mathbb R^d$ be a function with $\ell_2$-sensitivity $\Delta_2(f) := \max\limits_{X \sim X^\prime} \| f(X) - f(X') \|_2$.
The Gaussian mechanism defined as follow satisfies $\mu$-Gaussian Differential Privacy (GDP): 
\[
\mathcal M(X) = f(X) + Z,
\qquad
Z_i \stackrel{\text{i.i.d.}}{\sim} \mathcal{N}\!\left(0, \frac{\Delta_2(f)^2}{\mu^2}\right).
\]
\end{theorem}

Another type of question in \catone{} is report noisy max\citep{dwork2014algorithmic} for private selection, we provide a general template in Algorithm~\ref{alg:rnm}. In particular, when instantiated with exponential noise, the algorithm is called the permute and flip \citep{mckenna2020permute, ding2021permute}. When instantiated with Gumbel noise, the algorithm is the famous exponential algorithm \citep{mcsherry2007mechanism}.

\begin{algorithm}[H]
\caption{\textsc{ReportNoisyMax}}
\label{alg:rnm}
\DontPrintSemicolon
\SetKwInOut{KwIn}{Input}
\SetKwInOut{KwOut}{Output}

\KwIn{Dataset $X$; score functions $\{u_j(\cdot)\}_{i=1}^m$ with sensitivity $\Delta$ (w.r.t. same neighboring relation); noise distribution $\mathcal{P}$}
\KwOut{Index $\hat{i} \in \{1,\dots,m\}$ of the selected item.}

\For{$i=1$ \KwTo $m$}{
  $s_i \leftarrow u_i(X)$\;
  
  Sample $\eta_i \sim \mathcal{P}$\;
  
  $\tilde{s}_i \leftarrow s_i + \eta_i$\;
}
$\hat{i} \leftarrow \arg\max\limits_{i \in \{1,\dots,m\}} \tilde{s}_i$\;

\Return{$\hat{i}$}\;
\end{algorithm}

Before stating the privacy guarantee of private selection mechanisms used in \catone{}, we introduce definition of noises:

\begin{definition}[Exponential Distribution]
A random variable $X$ is said to follow an exponential distribution with parameter
$\lambda > 0$, denoted by $X \sim \mathrm{Exp}(\lambda)$, if it has probability density function
\[
f_X(x) =
\begin{cases}
\frac{1}{\lambda} e^{- x/\lambda}, & x \ge 0, \\
0, & x < 0.
\end{cases}
\]
\end{definition}

\begin{definition}[Gumbel Distribution]
A random variable $X$ is said to follow $\mathrm{Gumbel}(\alpha)$ distribution if it has probability density function
\[
f_X(x)
=
\frac{1}{\alpha}
\exp\!\left(
-\frac{x}{\alpha}
-
\exp\!\left(-\frac{x}{\alpha}\right)
\right),
\qquad x \in \mathbb R.
\]
\end{definition}

\begin{theorem}[Privacy guarantee of report noisy max with Laplace noise \citep{dwork2014algorithmic}] 
Suppose noise distribution $\mathcal{P}$ follows $\mathrm{Laplace}\left(\frac{2\Delta}{\varepsilon}\right)$, then Algorithm~\ref{alg:rnm} satisfies $\varepsilon$-DP 
\end{theorem}

\begin{theorem}[Privacy guarantee of report noisy max with Exponential noise \citep{ding2021permute}] 
Suppose noise distribution $\mathcal{P}$ follows $\mathrm{Exp}\left(\frac{2\Delta}{\varepsilon}\right)$, then Algorithm~\ref{alg:rnm} satisfies $\varepsilon$-DP. 
\end{theorem}

\begin{theorem}[Privacy guarantee of report noisy max with Gumbel noise \citep{ding2021permute}]\label{thm:laplaceRNM} 
Suppose noise distribution $\mathcal{P}$ follows $\mathrm{Gumbel}\left(\frac{2\Delta}{\varepsilon}\right)$, then Algorithm~\ref{alg:rnm} satisfies $\varepsilon$-DP. 
\end{theorem}

\subsection{References for \catwo{} questions}

In addition to DP textbook references \citep{dwork2014algorithmic, vadhan2017complexity}, we include in Table~\ref{tab:subject-topic-reference} the research papers used to construct and justify the \catwo{} questions. 

\begin{table}[H]
\centering
\small
\caption{Reference for Category 2 questions (grouped by subject and topic).}
\label{tab:subject-topic-reference}
\begin{tabular}{llp{0.5\linewidth}}
\toprule
Subject & Topic & Reference \\
\midrule
\multirow{5}{*}{Accounting and Composition} & Accounting & \citep{arxiv170207476, dong2022gaussian, arxiv200407223, ding2021permute, liu2022identification, harrison2025exact, dong2020optimal, durfee2019practical} \\
 & Amplification by Subsampling & \citep{arxiv180701647} \\
 & Parallel Composition & \citep{McSherry_2010} \\
 & SVT & \citep{lyu2017understanding} \\
 & Sequential or Adaptive Composition & \citep{bun2016concentrated, arxiv170207476, arxiv200400010, arxiv200407223} \\
\midrule
\multirow{8}{*}{DP-ML} & DP-GD & \citep{arxiv14057085, dong2022gaussian} \\
 & DP-OLS & \citep{pmlr-v70-sheffet17a} \\
 & Hyperparameter-tuning & \citep{arxiv211003620} \\
 & PATE & \citep{arxiv180208908} \\
 & dp-adam & \citep{arxiv211005679} \\
 & dp-sgd & \citep{Abadi_2016, arxiv220413650, rehn2025optimal} \\
 & objective perturbation & \citep{arxiv190809970, pmlr-v23-kifer12, chaudhuri2011differentially} \\
 & output perturbation & \citep{lowy2021output} \\
\midrule
\multirow{3}{*}{DP-statistics} & DP selection & \citep{arxiv190913830, dong2020optimal} \\
 & mean estimation & \citep{arxiv231206658} \\
 & quantile & \citep{pmlr-v162-kaplan22a} \\
\midrule
\multirow{2}{*}{Data-Adaptive} & PTR & \citep{arxiv230100301, dwork2009differential} \\
 & Smooth\_sensitivity & \citep{arxiv190602830} \\
\bottomrule
\end{tabular}

\end{table}

\section{Taxonomy of Error Patterns for Negative Questions in \catwo{}}\label{appendix:error_taxonomy}

Table~\ref{tab:error-taxonomy-dist} summarizes the error taxonomy for questions with negative labels in \catwo{}.

\begin{table}[H]
\centering
\small
\renewcommand{\arraystretch}{1.5}
\begin{threeparttable}
\caption{Error taxonomy in \catwo{}}
\label{tab:error-taxonomy-dist}
\begin{tabularx}{\textwidth}{p{0.28\textwidth} X c c}
\toprule
\textbf{Error pattern} & \textbf{Core characterization}  \\
\midrule

Overclaiming guarantee
& Starts from a valid result but claims a strictly stronger privacy guarantee than what the theorem or mechanism actually supports.
 \\

Broken implementation step
& The overall algorithm looks standard, but a key implementation detail is altered in a way that breaks privacy.
\\

Noise miscalibration
& Noise is added, but not at the level required by the true sensitivity or correct accounting.
 \\

Assumption mismatch
& The claim changes the assumptions, adjacency relation, norm bound, or sampling model while keeping the statement superficially similar.
\\

non-privatized data-dependent component
& A control-flow, hyperparameter, initialization, or stopping rule depends on private data without being privatized.
 \\

Public/private boundary mistake
& Quantities such as dataset size $n$ or $N$ are treated as public when they are private under the stated neighboring relation.
\\

Conceptual impossibility
& The claim is false at the conceptual level, such as invalid conversions between DP notions or privacy guarantees that are impossible in general.
 \\

Natural mistake
& The error originates in the literature
\\

\bottomrule
\end{tabularx}
\end{threeparttable}
\end{table}

\section{Experiment Details}\label{apx:exp_setup}

\subsection{Model Details}\label{apx:model}
We provide version information of models in the following table:
\begin{table}[H]
\centering
\fontsize{7}{10}\selectfont  %
\caption{Version information of Models}
\label{tab:models}
\begin{tabular}{lcccc}
\toprule[1.5pt]
Model & Provider & Version/Link/Release-date \\
\midrule[0.8pt]
\claudesonnet{} & Anthropic &  September 29, 2025\\
\claudeopus{} & Anthropic & November 24, 2025\\
\gemmin{} & Google DeepMind & June 17, 2025\\
\gemhigh{} & Google DeepMind & November 18, 2025\\ 
\gemthro{} & Google DeepMind & Feburary 19, 2026 \\
\gpthigh{} & OpenAI & August 7, 2025 \\
\gptmin{} & OpenAI &  August 7, 2025 \\
\gptffh{} & OpenAI & March 5, 2026 \\
\gptfvh{} & OpenAI & April 24, 2026 \\
\midrule[0.1pt]
\dpskrone{} & Open-sourced & \url{https://huggingface.co/deepseek-ai/DeepSeek-R1}\\
\dpskchat{} & Open-sourced & \url{https://huggingface.co/deepseek-ai/DeepSeek-V3.1} \\
\qweninst{} & Open-sourced & \url{https://huggingface.co/Qwen/Qwen3-30B-A3B-Instruct-2507} \\
\qwenthink{} & Open-sourced & \url{https://huggingface.co/Qwen/Qwen3-30B-A3B-Thinking-2507}\\
\goeprove{} & Open-sourced & \url{https://huggingface.co/Goedel-LM/Goedel-Prover-V2-32B}\\
\bottomrule[1.5pt]
\end{tabular}
\end{table}

\subsection{Experiment details for evaluation of \catone{} and \catwo{}}

For each question, we repeat the experiment with five random seeds. The prompt we use is as follows:

\begin{tcolorbox}[
    colback=gray!5,
    colframe=cyan!75!black,
    title=QA prompt for testing \catone{} and \catwo{},
    fonttitle=\bfseries
]
\{Question\} Please provide some explanations or deriviations first and then provide your final answer `yes` or `no` inside a Latex boxed format `\texttt{\textbackslash\textbackslash boxed\{\}}` 
\end{tcolorbox}

\subsection{Experiment details for Augmented QA prompt}

\subsubsection{QA prompt augmented with relevant theorems}
\begin{tcolorbox}[
    colback=gray!5,
    colframe=cyan!75!black,
    title=QA prompt augmented with relevant theorems,
    fonttitle=\bfseries
]

\{Question description\}

You may use the following theorems or definitions if applicable. 

[THEOREM]
\{Theorem\}
[/THEOREM]

Please provide explanations or derivations first. If you use any theorem or definition, cite it by its ID in the reasoning. The very last line of your response must be exactly `\texttt{\textbackslash\textbackslash boxed\{yes\}}` or `\texttt{\textbackslash\textbackslash boxed\{no\}}`. Do not include citations or any other text on the last line.
\end{tcolorbox}

\subsubsection{QA prompt with one-shot proof template}\label{apx:one-shot-temp}

\begin{tcolorbox}[
    colback=gray!5,
    colframe=cyan!75!black,
    title=QA template for one-shot reasoning on LaplaceRNM task,
    fonttitle=\bfseries
]

Below is an example of the question.\\

Example Question:\\
For each item $j\in F$, define a score $s_j(D)$ with sensitivity 1. Consider the mechanism that samples independent noise $Z_j$ ~ Laplace(2/$\epsilon$) for each $j$ and outputs
$\argmax_j (s_j(D) + Z_j)$.\\

Claim: This mechanism satisfies $\epsilon$-differential privacy.\\

Now answer the following question.\\

\{Question description\} \\

Please provide some explanations or deriviations first and then provide your final answer `yes` or `no` inside a Latex boxed format `\texttt{\textbackslash\textbackslash boxed\{\}}`.

\end{tcolorbox}

\newpage
\subsection{Experiment details for analyzing LLMs' responses}

For the judge LLM, we use GPT-5.4 with high reasoning effort. 

\subsubsection{Prompt for judging LLMs' response}\label{apx:response_judge}

\begin{tcolorbox}[
    colback=gray!5,
    colframe=cyan!75!black,
    title=Prompt for judging LLM response,
    fonttitle=\bfseries
]

You are an expert evaluator of differential privacy reasoning.\\

Evaluate whether the model response gives a mathematically reasonable argument for the stated problem.\\
Do not require a fully formal proof. Minor omissions are acceptable if the main idea is correct and the conclusion is supported.\\

Use the ground-truth label only to check whether the response's final conclusion matches the benchmark.\\
Hints are optional reference information. The response does not need to use the hints or follow the same solution path. A response can be correct if it gives a different mathematically valid argument.\\

Classify the response into one category:\\

- "correct": sound and sufficient reasoning.\\
- "minor\_gap": mostly correct with small omissions or imprecision.\\
- "correct\_direction\_fixable": right high-level idea, but missing or mistaken key details that are locally fixable.\\
- "fatal\_error": core mathematical error that invalidates the reasoning.\\
- "unclear\_or\_insufficient": too vague or lacks substantive reasoning.\\

Set reasoning\_correct as:
- "yes" for correct or minor\_gap\\
- "partial" for correct\_direction\_fixable\\
- "no" for fatal\_error or unclear\_or\_insufficient\\

Return exactly one JSON object:\\
\{\\
  "reasoning\_correct": "yes" or "partial" or "no",\\
  "error\_category": "correct" or "minor\_gap" or "correct\_direction\_fixable" or "fatal\_error" or "unclear\_or\_insufficient",\\
  "error\_summary": "Briefly summarize the main issue, or state why the reasoning is acceptable.",\\
  "detailed\_explanation": "2-5 sentences explaining the assessment. Do not penalize the response for not using the hints if its alternative reasoning is valid." \\
\}
\\

Problem:\\
\{problem\}
\\

Ground-truth answer:\\
\{label\}
\\

Hints, if available:\\
\{hints\}
\\

Model response:\\
\{response\}

\end{tcolorbox}

\subsubsection{Description of Fine-Grained Reasoning Assessment Labels}

\begin{table}[H]
\centering
\begin{tabular}{lp{7cm}c}
\toprule
\textbf{Error Category} & \textbf{Description}  \\
\midrule
    \texttt{correct} & sound and sufficient reasoning  \\
    \texttt{minor\_gap} & mostly correct with small omissions or imprecision  \\
    \texttt{correct\_direction\_fixable} & right high-level idea, but missing or mistaken key details that are fixable  \\
    \texttt{fatal\_error} & core mathematical error that invalidates the reasoning \\
    \texttt{unclear\_or\_insufficient} & too vague or lacks substantive reasoning \\
\bottomrule
\end{tabular}
\caption{Description of reasoning assessment labels}
\label{tab:softlabels}
\end{table}

\subsubsection{Calculation for plausibility and soundness scores}\label{sec:reasoningmetric}
Define shorthand for labels: $C$ : \texttt{correct}, $M$ : \texttt{minor\_gap}, $F$ : \texttt{correct\_direction\_fixable}. The Soundness and Plausibility are calculated as follow: 
\begin{equation}
    \begin{aligned}
        \mathrm{Soundness}&=\frac{
\#\{\text{answer correct} \land C\} + \#\{\text{answer correct} \land M\}
}{
\#\{\text{answer correct}\}
} \\
        \mathrm{Plausibility}&=\frac{
\#\{\text{answer correct} \land C\} + \#\{\text{answer correct} \land M\} + \#\{\text{answer correct} \land F\} 
}{
\#\{\text{answer correct}\}
}
    \end{aligned}
\end{equation}

\section{Accuracy per topic for \catwo{}}

\begin{table}[h]
\centering
\fontsize{8}{10}\selectfont  %
\caption{The performance of \gpthigh{} and \gemhigh{} on \catwo{} by topics. We sort the topics by the mean accuracy of two models.}
\label{tab:cat2_complete}
\begin{tabular}{lrrrrr}
\toprule[1.5pt]
Topic & Gemini-3-Pro & Gemini-3.1-Pro & GPT-5.4-High & GPT-5.5-High & Avg Accuracy \\
\midrule
DP Adam & 1.00 & 1.00 & 1.00 & 1.00 & 1.00 \\
Quantile Estimation & 1.00 & 1.00 & 1.00 & 1.00 & 1.00 \\
Privacy Accounting & 0.92 & 1.00 & 1.00 & 0.98 & 0.97 \\
Private Selection & 0.80 & 1.00 & 1.00 & 1.00 & 0.95 \\
Privacy Amplification via Subsampling & 1.00 & 1.00 & 0.96 & 1.00 & 0.94 \\
Sparse Vector Technique & 0.92 & 0.95 & 1.00 & 1.00 & 0.94 \\
DP Gradient Descent & 0.92 & 0.92 & 0.92 & 0.88 & 0.92 \\
Sequential/Adaptive Composition & 0.91 & 0.97 & 0.94 & 0.97 & 0.90 \\
Mean Estimation & 0.87 & 0.85 & 0.87 & 0.87 & 0.86 \\
DP Linear Regression & 0.80 & 0.93 & 0.80 & 0.83 & 0.85 \\
Objective Perturbation & 0.80 & 0.80 & 0.70 & 0.90 & 0.80 \\
Parallel Composition & 0.71 & 0.84 & 0.77 & 0.86 & 0.80 \\
DP Stochastic Gradient Descent & 0.78 & 0.78 & 0.82 & 0.76 & 0.79 \\
DP Hyperparameter Optimization & 0.60 & 0.70 & 0.67 & 1.00 & 0.75 \\
Propose–Test–Release (PTR) & 0.70 & 0.75 & 0.85 & 0.75 & 0.74 \\
Output Perturbation & 0.33 & 0.87 & 0.40 & 0.67 & 0.57 \\
Private Aggregation of Teacher Ensembles (PATE) & 0.67 & 0.60 & 0.40 & 0.60 & 0.52 \\
Smooth Sensitivity & 0.33 & 0.33 & 0.33 & 0.33 & 0.33 \\
\bottomrule[1.5pt]
\end{tabular}
\end{table}

\section{Benchmark Examples}
\label{appendix:example}
\paragraph{Example of the positive question in category 2:} Suppose that for all $z\in \mathcal{Z}$, $\ell(\cdot, z)$ is twice-differentiable, and the rank of its Hessian $\nabla^2\ell(\mathbf{w}, z)$ at any $\mathbf{w}\in \mathcal{W}$ is at most 1. Also assume that  the smoothness parameter satisfies $\beta \leq \varepsilon n\lambda$. Is the following algorithm$(\varepsilon, \delta)$-differentially private?

\begin{algorithm}[H]
\caption{$\mathcal{A}_{\text{ObjP}}$: Objective Perturbation}
\KwIn{Private dataset $S = (z_1, \ldots, z_n) \in \mathcal{Z}^n$, 
$L$-Lipschitz, $\beta$-smooth, convex loss function $\ell$, convex set 
$\mathcal{W} \subseteq \mathbb{R}^d$, privacy parameters 
$\varepsilon \le 1$, $\delta \le 1/n^2$, regularization parameter $\lambda$.}

\BlankLine
\textbf{1:} Sample $\mathbf{G} \sim \mathcal{N}(\mathbf{0}, \sigma^2 \mathbf{I}_d)$, 
where $\sigma^2 = \dfrac{10 L^2 \log(1/\delta)}{\varepsilon^2}$\;

\textbf{2:} \textbf{return} 
$\hat{\mathbf{w}} = 
\arg\min_{\mathbf{w} \in \mathcal{W}} 
\hat{\mathcal{L}}(\mathbf{w}; S) 
+ \dfrac{\langle \mathbf{G}, \mathbf{w} \rangle}{n} 
+ \lambda \|\mathbf{w}\|^2$, 
where $\hat{\mathcal{L}}(\mathbf{w}; S) 
\triangleq \dfrac{1}{n} \sum_{i=1}^{n} \ell(\mathbf{w}, z_i)$\;

\end{algorithm}

\paragraph{Example of the negative question (\textit{assumption perturbation}) in category 2:}
Suppose that for all $z\in \mathcal{Z}$, $\ell(\cdot, z)$ is twice-differentiable, and the rank of its Hessian $\nabla^2\ell(\mathbf{w}, z)$ at any $\mathbf{w}\in \mathcal{W}$ is at most 1. Also assume that  the smoothness parameter satisfies $\beta \leq \varepsilon n\lambda$. Is the following algorithm$(\varepsilon, \delta)$-differentially private?

\begin{algorithm}[H]
\caption{$\mathcal{A}_{\text{ObjP}}$: Objective Perturbation}
\KwIn{Private dataset $S = (z_1, \ldots, z_n) \in \mathcal{Z}^n$, 
$L$-Lipschitz, $\beta$-smooth, {\color{red}\cancel{convex}} loss function $\ell$, privacy parameters 
$\varepsilon \le 1$, $\delta \le 1/n^2$, regularization parameter $\lambda$.}

\BlankLine
\textbf{1:} Sample $\mathbf{G} \sim \mathcal{N}(\mathbf{0}, \sigma^2 \mathbf{I}_d)$, 
where $\sigma^2 = \dfrac{10 L^2 \log(1/\delta)}{\varepsilon^2}$\;

\textbf{2:} \textbf{return} 
$\hat{\mathbf{w}} = 
\arg\min_{\mathbf{w} \in \mathcal{W}} 
\hat{\mathcal{L}}(\mathbf{w}; S) 
+ \dfrac{\langle \mathbf{G}, \mathbf{w} \rangle}{n} 
+ \lambda \|\mathbf{w}\|^2$, 
where $\hat{\mathcal{L}}(\mathbf{w}; S) 
\triangleq \dfrac{1}{n} \sum_{i=1}^{n} \ell(\mathbf{w}, z_i)$\;

\end{algorithm}

\paragraph{Example of the negative question (\textit{algorithm perturbation}) in category 2:}
Suppose that for all $z\in \mathcal{Z}$, $\ell(\cdot, z)$ is twice-differentiable, and the rank of its Hessian $\nabla^2\ell(\mathbf{w}, z)$ at any $\mathbf{w}\in \mathcal{W}$ is at most 1. Also assume that  the smoothness parameter satisfies $\beta \leq \varepsilon n\lambda$. Is the following algorithm$(\varepsilon, \delta)$-differentially private?

\begin{algorithm}[H]
\caption{$\mathcal{A}_{\text{ObjP}}$: Objective Perturbation}
\KwIn{Private dataset $S = (z_1, \ldots, z_n) \in \mathcal{Z}^n$, 
$L$-Lipschitz, $\beta$-smooth, convex loss function $\ell$, convex set 
$\mathcal{W} \subseteq \mathbb{R}^d$, privacy parameters 
$\varepsilon \le 1$, $\delta \le 1/n^2$, regularization parameter $\lambda$.}

\BlankLine
\textbf{1:} Sample $\mathbf{G} \sim \mathcal{N}(\mathbf{0}, \sigma^2 \mathbf{I}_d)$, 
where $\sigma^2 = \dfrac{10 L^2 \log(1/\delta)}{\varepsilon^2}$\;

\textbf{2:} \textbf{return} 
$\hat{\mathbf{w}} = 
\arg\min_{\mathbf{w} \in \mathcal{W}} 
\hat{\mathcal{L}}(\mathbf{w}; S) 
+ \dfrac{\langle \mathbf{G}, \mathbf{w} \rangle}{n} 
+ {\color{red}\frac{\lambda}{2} \|\mathbf{w}\|^2}$, 
where $\hat{\mathcal{L}}(\mathbf{w}; S) 
\triangleq \dfrac{1}{n} \sum_{i=1}^{n} \ell(\mathbf{w}, z_i)$\;

\end{algorithm}

\paragraph{Example of the negative question (\textit{conclusion perturbation}) in category 2:} Suppose that for all $z\in \mathcal{Z}$, $\ell(\cdot, z)$ is twice-differentiable, and the rank of its Hessian $\nabla^2\ell(\mathbf{w}, z)$ at any $\mathbf{w}\in \mathcal{W}$ is at most 1. Also assume that  the smoothness parameter satisfies $\beta \leq \varepsilon n\lambda$. Is the following algorithm$({\color{red}\varepsilon/2}, \delta)$-differentially private?

\begin{algorithm}[H]
\caption{$\mathcal{A}_{\text{ObjP}}$: Objective Perturbation}
\KwIn{Private dataset $S = (z_1, \ldots, z_n) \in \mathcal{Z}^n$, 
$L$-Lipschitz, $\beta$-smooth, convex loss function $\ell$, convex set 
$\mathcal{W} \subseteq \mathbb{R}^d$, privacy parameters 
$\varepsilon \le 1$, $\delta \le 1/n^2$, regularization parameter $\lambda$.}

\BlankLine
\textbf{1:} Sample $\mathbf{G} \sim \mathcal{N}(\mathbf{0}, \sigma^2 \mathbf{I}_d)$, 
where $\sigma^2 = \dfrac{10 L^2 \log(1/\delta)}{\varepsilon^2}$\;

\textbf{2:} \textbf{return} 
$\hat{\mathbf{w}} = 
\arg\min_{\mathbf{w} \in \mathcal{W}} 
\hat{\mathcal{L}}(\mathbf{w}; S) 
+ \dfrac{\langle \mathbf{G}, \mathbf{w} \rangle}{n} 
+ \lambda \|\mathbf{w}\|^2$, 
where $\hat{\mathcal{L}}(\mathbf{w}; S) 
\triangleq \dfrac{1}{n} \sum_{i=1}^{n} \ell(\mathbf{w}, z_i)$\;

\end{algorithm}

\newpage
\section{Examples of Fine-Grained Error Analysis with GPT-5.4-High as Judge}\label{apx:llm-as-judge}
We include examples of judge results for responses generated by GPT-5-High. \textbf{GT} denotes the ground truth label, \textbf{Pred} denotes the label predicted by the response LLM.

\lstset{
  basicstyle=\ttfamily\scriptsize,
  breaklines=true,
  breakatwhitespace=false,
  columns=flexible,
  keepspaces=true,
  frame=single,
  framerule=0.4pt,
  rulecolor=\color{gray!50},
  backgroundcolor=\color{gray!5},
  xleftmargin=4pt,xrightmargin=4pt,
  aboveskip=4pt,belowskip=2pt,
}
\setlength{\parindent}{0pt}

\thispagestyle{empty}
\pagestyle{plain}
\vspace{6pt}
\begin{tcolorbox}[colback=catcorrect,colframe=catcorrect,boxrule=0pt,arc=3pt,left=6pt,right=6pt,top=5pt,bottom=5pt]\color{white}\begin{tabular}{@{}p{1.1cm}p{2.4cm}p{2.4cm}p{7.4cm}@{}}\textbf{Q\,18} & GT:~\textbf{Yes (1)} & Pred:~\textbf{Yes (1)} & \textbf{Correct}~~{\color{catcorrect}\checkmark\ correct}\end{tabular}\end{tcolorbox}
\begin{tcolorbox}[colback=gray!7,colframe=gray!35,boxrule=0.5pt,arc=2pt,left=8pt,right=8pt,top=5pt,bottom=5pt]\noindent{\small\color{gray}\textit{DP-statistics $\triangleright$ DP selection: expoMech}}\\[4pt]
The pseudocode below implements the exponential mechanism. Does this implementation ensure $\varepsilon$-differential privacy under the add/remove neighboring relationship?

\BlankLine
\end{tcolorbox}
\vspace{2pt}
\begin{algorithm}[H]
\caption{Exponential Mechanism}
\DontPrintSemicolon
\SetNoFillComment
\SetKwInOut{Input}{Input}
\SetKwInOut{Output}{Output}

\Input{
Dataset $D = \{z_i\}_{i=1}^n$;\,
candidate set $Y$;\,
utility (score) function $q(D, y)=\sum_{i=1}^n \mathbf{1}\{z_i=y\}$;\,
privacy parameter $\varepsilon$
}

\Output{
Selected item $\hat{y} \in Y$ sampled by the exponential mechanism
}

\BlankLine
\For{$y \in Y$}{
    $p_y \leftarrow 
    \frac{\exp\left(\varepsilon\, q(D, y)\right)}
         {\sum_{y' \in Y} \exp\left(\varepsilon\, q(D, y')\right)}$
}

Sample $\hat{y} \sim \mathrm{Multinomial}(1; \{p_y\}_{y \in Y})$\;
\Return{$\hat{y}$}
\end{algorithm}
\vspace{2pt}
\vspace{2pt}
{\sffamily\small\textbf{Comments from metadata}}\par\vspace{2pt}
\begin{tcolorbox}[colback=blue!4,colframe=blue!30,boxrule=0.5pt,arc=2pt,left=8pt,right=8pt,top=5pt,bottom=5pt]
{\small For a monotone query $q$, the exponential mechanism with $p_y \propto \exp\!\left(\frac{\varepsilon\, q}{\Delta_q}\right)$ satisfies $\varepsilon$-DP \citep{dong2020optimal}.}
\end{tcolorbox}
{\sffamily\small\textbf{Model Response}~{\color{gray}(gpt-5-high)}}\par\vspace{2pt}
\begin{lstlisting}
Short answer with context

- The exponential mechanism with generic utility u ensures epsilon-DP when it samples with probability proportional to exp((epsilon u(D,y))/(2Deltau)), where Deltau is the global sensitivity. The factor 1/2 is needed in general because both the numerator term and the normalizing constant can each contribute a factor exp(epsilon/2) in the privacy ratio.

- In your pseudocode, the utility is the count q(D,y) = sumi 1{zi = y}. Under add/remove adjacency, this utility is monotone with respect to adding a record: if D' = D union {y*}, then q(D',y) = q(D,y) + 1{y = y*} (no utility decreases). This monotonicity allows a tighter analysis that eliminates the extra factor 1/2.

Derivation for the add/remove case

Let D' = D union {y*}. Define ZD = sumy exp(epsilon q(D,y)) and ZD' = sumy exp(epsilon q(D',y)).

- Because q(D',y) = q(D,y) + 1{y=y*}, we have
  ZD' = ZD + (e^epsilon - 1) exp(epsilon q(D,y*)) = ZD * [1 + (e^epsilon - 1) wD(y*)],
  where wD(y*) = exp(epsilon q(D,y*))/ZD in [0,1]. Hence 1 <= ZD'/ZD <= e^epsilon.

- For y = y*, the ratio of output probabilities is
  pD(y*)/pD'(y*) = e^{-epsilon} * (ZD'/ZD) in [e^{-epsilon}, 1] subseteq [e^{-epsilon}, e^{epsilon}].

- For y != y*, the ratio is
  pD(y)/pD'(y) = ZD'/ZD in [1, e^{epsilon}] subseteq [e^{-epsilon}, e^{epsilon}].

Thus for every y, e^{-epsilon} <= pD(y)/pD'(y) <= e^{epsilon}. Summing over any subset S subseteq Y preserves the bound, so the mechanism satisfies epsilon-differential privacy under add/remove adjacency.

Remarks
- For substitution (replace-one) adjacency or for non-monotone utilities, the same code would only be 2epsilon-DP; to get epsilon-DP in that generality you need the standard scaling exp((epsilon q(D,y))/(2Deltaq)).
- The argument assumes the candidate set Y is fixed (data-independent).

Final answer:
\boxed{yes}
\end{lstlisting}

{\sffamily\small\textbf{Judge Assessment}~{\color{gray}(gpt-5.4)}}\par\vspace{2pt}
\begin{tcolorbox}[colback=catcorrectbg,colframe=catcorrect,boxrule=0.5pt,arc=2pt,left=8pt,right=8pt,top=5pt,bottom=5pt]
{\color{catcorrect}\textbf{Reasoning correct}}\hfill\textbf{Category:}~{\color{catcorrect}\textbf{Correct}}\\[3pt]
\textbf{Summary:}~The response gives a valid monotonicity-based privacy argument showing that the unscaled exponential mechanism is epsilon-DP for this count utility under add/remove adjacency.\\[3pt]
\textbf{Detail:}~The key observation is correct: when one record is added, only the score of the matching item increases by 1, so the utility is monotone and the partition function changes by at most a factor of e\textasciicircum{}epsilon. Using this, the response correctly bounds the pointwise output-probability ratios for both the affected y* and all other outputs by e\textasciicircum{}epsilon (and below by e\textasciicircum{}\{-epsilon\}), which is sufficient for epsilon-DP. The note that the usual 1/2 factor is needed only in the general non-monotone or replace-one setting is also mathematically appropriate.
\end{tcolorbox}
\vspace{4pt}\noindent\rule{\linewidth}{0.4pt}

\vspace{6pt}
\begin{tcolorbox}[colback=catcorrect,colframe=catcorrect,boxrule=0pt,arc=3pt,left=6pt,right=6pt,top=5pt,bottom=5pt]\color{white}\begin{tabular}{@{}p{1.1cm}p{2.4cm}p{2.4cm}p{7.4cm}@{}}\textbf{Q\,35} & GT:~\textbf{No (0)} & Pred:~\textbf{No (0)} & \textbf{Correct}~~{\color{catcorrect}\checkmark\ correct}\end{tabular}\end{tcolorbox}
\begin{tcolorbox}[colback=gray!7,colframe=gray!35,boxrule=0.5pt,arc=2pt,left=8pt,right=8pt,top=5pt,bottom=5pt]\noindent{\small\color{gray}\textit{DP-ML $\triangleright$ DP-GD}}\\[4pt]
The following is an implementation of differentially private gradient descent (DP-GD). Does this implementation satisfy $\mu$-Gaussian differential privacy under the add/remove neighboring relationship (assume number of samples $n$ is public information)?
\end{tcolorbox}
\vspace{2pt}
\begin{algorithm}[H]
\caption{DP-GD with Gaussian Noise }
 
\DontPrintSemicolon
\SetNoFillComment
\SetKwInOut{Input}{Input}\SetKwInOut{Output}{Output}
\Input{
Dataset $D=\{z_i\}_{i=1}^n$;\,
loss $\ell(\theta;z)$;\,
initial parameter $\theta_0\in\mathbb{R}^d$;\,
iterations $T$;\,
step sizes $\{\eta_t\}_{t=0}^{T-1}$;\,
clipping norm $C>0$;\,
target gaussian DP $\mu$;\,
Stopping threshold $\tau$
}
\Output{Final parameter $\theta_T$  }
\BlankLine

\For{$t=0,\dots,T-1$}{
  \For{$i=1,\dots,n$}{
    $g_i \leftarrow \nabla_\theta \ell(\theta_t; z_i)$\;
    $\hat g_i \leftarrow  g_i \cdot \min\!\bigl(1,\, \tfrac{C}{\|g_i\|_2}\bigr)$\;
  }
  $\tilde g_t \leftarrow \frac{1}{n}\left(\sum_{i=1}^n \hat g_i + \frac{C\sqrt{2T}}{\mu}\xi_t \right),\quad \xi_t \sim \mathcal{N}(0, I_d)$\; 
  $\theta_{t+1} \leftarrow \theta_t \;-\; \eta_t \,\tilde g_t$\;
  \tcp{running above-threshold}
  \If{$\sum_{i=1}^{n}\ell(\theta_{t+1};z_i) + \frac{C\sqrt{2T}}{\mu}\xi_t \geq \tau$}{
        \textbf{break} 
  }
  
}

\Return{$\theta_{t+1}$}
\end{algorithm}
\vspace{2pt}
\vspace{2pt}
{\sffamily\small\textbf{Comments from metadata}}\par\vspace{2pt}
\begin{tcolorbox}[colback=blue!4,colframe=blue!30,boxrule=0.5pt,arc=2pt,left=8pt,right=8pt,top=5pt,bottom=5pt]
{\small The noise added to the threshold is not properly calibrated to its sensitivity.}
\end{tcolorbox}
{\sffamily\small\textbf{Model Response}~{\color{gray}(gpt-5-high)}}\par\vspace{2pt}
\begin{lstlisting}
Short answer: No, because the early-stopping "above-threshold" check is not privatized correctly.

Explanation/derivation:

- GDP basics. A Gaussian mechanism that outputs f(D) + N(0, sigma^2 I) with L2-sensitivity Delta under add/remove adjacency is exactly mu-GDP with mu = Delta/sigma. For a sequence of (possibly adaptive) Gaussian mechanisms with mu1, ..., muT, composition in GDP gives mutotal = sqrt(sumt mut^2).

- Gradient step (ignoring the stopping rule). The per-iteration query is the averaged, per-example-clipped gradient:
  f_t(D) = (1/n) sumi hat g_i, with ||hat g_i||2 <= C.
  Under add/remove adjacency, the L2-sensitivity is Deltagrad = C/n. The code adds Gaussian noise with std to the average equal to sigmagrad = (Csqrt(2T))/(mu n). Hence the per-iteration GDP parameter is mut = Deltagrad/sigmagrad = mu/sqrt(2T). Composing T iterations gives mugrad,total = sqrt(T*(mu^2/(2T))) = mu/sqrt2 <= mu. So the gradient part alone would be at least as private as mu-GDP (indeed, slightly stronger).

- Early-stopping "above-threshold" step breaks privacy:
  1) The statistic being tested is S_t(D) = sumi l(theta_{t+1}; z_i). Under add/remove adjacency, the sensitivity of S_t is DeltaS = supz |l(theta; z)|, which is unbounded unless the loss is explicitly bounded or clipped. The code does not bound or clip the loss.
  2) The noise scale used for the threshold test is tied to the gradient clipping constant C, which has no relation to DeltaS. Calibrating to C does not protect the sum of losses.
  3) The same noise sample xi_t is (apparently) reused for both the vector-valued gradient noise and the scalar threshold test, which is both dimensionally inconsistent and, even if made consistent, undermines clean composition accounting; at minimum one needs fresh, independently drawn noise for the test, calibrated to the sensitivity of S_t and budgeted in composition.

Because the stopping condition depends directly on the (non-privatized) sum of losses, the stopping time can leak information about the presence/absence of a point. Since the final output depends on whether the loop stops, this additional data access is not protected by "post-processing" closure (post-processing only applies to functions of DP outputs, not fresh queries to the data). Therefore, as written, the overall algorithm does not satisfy mu-GDP.

If the above-threshold step were removed, or replaced by a properly privatized AboveThreshold mechanism with bounded loss sensitivity and independently calibrated noise accounted for in composition (ensuring mugrad,total^2 + muthresh,total^2 <= mu^2), then one could achieve mu-GDP. But in its current form, it does not.

\boxed{No}
\end{lstlisting}

{\sffamily\small\textbf{Judge Assessment}~{\color{gray}(gpt-5.4)}}\par\vspace{2pt}
\begin{tcolorbox}[colback=catcorrectbg,colframe=catcorrect,boxrule=0.5pt,arc=2pt,left=8pt,right=8pt,top=5pt,bottom=5pt]
{\color{catcorrect}\textbf{Reasoning correct}}\hfill\textbf{Category:}~{\color{catcorrect}\textbf{Correct}}\\[3pt]
\textbf{Summary:}~The response correctly identifies that the gradient updates alone can be made mu-GDP under composition, but the early-stopping loss-threshold query is not properly privatized because its noise is not calibrated to that query's sensitivity.\\[3pt]
\textbf{Detail:}~The response gives a mathematically reasonable privacy analysis of the clipped-gradient Gaussian mechanism: under add/remove adjacency the averaged clipped gradient has sensitivity C/n, and the stated noise scale yields a per-step GDP parameter that composes to at most mu for the gradient part. It then correctly notes that the stopping rule performs an additional data-dependent query, namely the sum of losses, and this is not protected by post-processing since it accesses the dataset anew. The key flaw is that the threshold-test noise is tied to the gradient clipping constant C rather than the sensitivity of the loss sum (which may even be unbounded without a bounded/clipped loss assumption), so the overall algorithm as written does not satisfy mu-GDP.
\end{tcolorbox}
\vspace{4pt}\noindent\rule{\linewidth}{0.4pt}

\vspace{6pt}
\begin{tcolorbox}[colback=catfatal,colframe=catfatal,boxrule=0pt,arc=3pt,left=6pt,right=6pt,top=5pt,bottom=5pt]\color{white}\begin{tabular}{@{}p{1.1cm}p{2.4cm}p{2.4cm}p{7.4cm}@{}}\textbf{Q\,92} & GT:~\textbf{No (0)} & Pred:~\textbf{Yes (1)} & \textbf{Fatal Error}~~{\color{catfatal}\(\times\)\ wrong}\end{tabular}\end{tcolorbox}
\begin{tcolorbox}[colback=gray!7,colframe=gray!35,boxrule=0.5pt,arc=2pt,left=8pt,right=8pt,top=5pt,bottom=5pt]\noindent{\small\color{gray}\textit{DP-ML $\triangleright$ output perturbation}}\\[4pt]
Let $\varepsilon > 0$, $\delta \in [0, \frac{1}{2})$. Let $F : \mathbb{R}^d \times \mathcal{X}^n \to \mathbb{R}$ be a function with $L_2$ sensitivity $\Delta_F$ and unique minimizer $w^*(X) = \arg\min_{w \in \mathbb{R}^d} F(w, X)$ for all $X \in \mathcal{X}^n$. The (conceptual) output perturbation algorithms $\mathcal{A}(X) = w_{\mathcal{A}}(X) := w^*(X) + z$ with noise given by density
$$
p_z(t) \propto 
\begin{cases}
\exp\!\left(-\dfrac{\varepsilon \|t\|_2}{\Delta_F}\right), & \text{if } \delta = 0,\\[1em]
\exp\!\left(-\dfrac{\varepsilon \|t\|_2}{\Delta_F (c_\delta + \sqrt{c_\delta^2 + \varepsilon})}\right), & \text{if } \delta \in (0, \tfrac{1}{2})
\end{cases}
$$
where $c_\delta := \sqrt{\log\!\left(\dfrac{2}{\sqrt{16\delta^2 + 1} - 1}\right)}$. Is $\mathcal{A}(X)$ $(\varepsilon, \delta)$-differentially private?
\end{tcolorbox}
\vspace{2pt}
{\sffamily\small\textbf{Comments from metadata}}\par\vspace{2pt}
\begin{tcolorbox}[colback=blue!4,colframe=blue!30,boxrule=0.5pt,arc=2pt,left=8pt,right=8pt,top=5pt,bottom=5pt]
{\small (1) Not necessarily: adding Laplace/Gaussian noise to $w^*(X)$ is $(\varepsilon,\delta)$-DP only if the noise is calibrated to the $L_2$-sensitivity of $w^*(X)$, which does not follow from assuming $F$ has sensitivity $\Delta_F$ without additional stability/regularity conditions; (2) the density for $\delta \in (0,1/2)$ is not correct.}
\end{tcolorbox}
{\sffamily\small\textbf{Model Response}~{\color{gray}(gpt-5-high)}}\par\vspace{2pt}
\begin{lstlisting}
Explanation/derivation:
- Let the "query" be q(X) := w*(X). Interpret DeltaF as the L2 global sensitivity of this query: for all neighboring datasets X, X', we have ||q(X) - q(X')||2 <= DeltaF.
- Consider additive noise Z with density galpha(z) propto exp(-alpha ||z||2), and output A(X) = q(X) + Z. For any y in Rd,
  log(p(y|X)/p(y|X')) = -alpha(||y - q(X)||2 - ||y - q(X')||2) <= alpha ||q(X) - q(X')||2 <= alpha DeltaF,
  where we used the triangle inequality and the fact the normalizing constant cancels. Hence A is (alpha DeltaF, 0)-DP.
- delta = 0 case: choosing alpha = epsilon/DeltaF gives (epsilon, 0)-DP (the K-norm/L2-Laplace mechanism).
- delta in (0, 1/2) case: the proposed density uses alpha = epsilon/[DeltaF(cdelta + sqrt(cdelta^2 + epsilon))], so the mechanism is (epsilon', 0)-DP with epsilon' = alpha DeltaF = epsilon/(cdelta + sqrt(cdelta^2 + epsilon)) <= epsilon. Pure DP with parameter epsilon' <= epsilon trivially implies (epsilon, delta)-DP.

Remark: If DeltaF is meant as the L2 sensitivity of q(X) = w*(X), the above holds. If instead DeltaF referred only to sensitivity of the scalar-valued objective F(*, X) while the argmin's sensitivity is unbounded, no finite-noise DP guarantee follows. Under the usual interpretation (DeltaF is the sensitivity of w*), the mechanism is indeed (epsilon, delta)-DP; in fact it is strictly stronger (pure DP) in both cases.

Final answer:
\boxed{\,\text{Yes}\,}
\end{lstlisting}

{\sffamily\small\textbf{Judge Assessment}~{\color{gray}(gpt-5.4)}}\par\vspace{2pt}
\begin{tcolorbox}[colback=catfatalbg,colframe=catfatal,boxrule=0.5pt,arc=2pt,left=8pt,right=8pt,top=5pt,bottom=5pt]
{\color{catfatal}\textbf{Reasoning incorrect}}\hfill\textbf{Category:}~{\color{catfatal}\textbf{Fatal Error}}\\[3pt]
\textbf{Summary:}~The response incorrectly treats Delta\_F as the L2 sensitivity of the argmin query w*(X), but the problem only assumes sensitivity of the objective F, which does not imply bounded sensitivity of w*(X).\\[3pt]
\textbf{Detail:}~The main proof would be valid for additive L2-Laplace noise if Delta\_F were a bound on ||w*(X)-w*(X')||\_2 for neighboring datasets. However, that is not what the problem states: sensitivity of F(w,X) does not by itself control the sensitivity of its minimizer, so the DP argument does not apply. The response even acknowledges this caveat but then answers "Yes" anyway, which is incompatible with the stated assumptions. It also does not address that the proposed delta>0 calibration is not the standard correct one.
\end{tcolorbox}
\vspace{4pt}\noindent\rule{\linewidth}{0.4pt}

\vspace{6pt}
\begin{tcolorbox}[colback=catminor,colframe=catminor,boxrule=0pt,arc=3pt,left=6pt,right=6pt,top=5pt,bottom=5pt]\color{white}\begin{tabular}{@{}p{1.1cm}p{2.4cm}p{2.4cm}p{7.4cm}@{}}\textbf{Q\,95} & GT:~\textbf{No (0)} & Pred:~\textbf{No (0)} & \textbf{Minor Gap}~~{\color{catcorrect}\checkmark\ correct}\end{tabular}\end{tcolorbox}
\begin{tcolorbox}[colback=gray!7,colframe=gray!35,boxrule=0.5pt,arc=2pt,left=8pt,right=8pt,top=5pt,bottom=5pt]\noindent{\small\color{gray}\textit{Accounting and Composition $\triangleright$ Parallel Composition}}\\[4pt]
We aim to release a histogram whose bins are $n$ data-independent but potentially non-disjoint intervals. 
If we add independent noise drawn from $\mathrm{Lap}(2/\varepsilon)$ to each bin, 
does the resulting mechanism satisfy $\varepsilon$-differential privacy under the replace-one neighboring relationship?
\end{tcolorbox}
\vspace{2pt}
{\sffamily\small\textbf{Comments from metadata}}\par\vspace{2pt}
\begin{tcolorbox}[colback=blue!4,colframe=blue!30,boxrule=0.5pt,arc=2pt,left=8pt,right=8pt,top=5pt,bottom=5pt]
{\small Since the histogram bins may be non-disjoint, replacing a single data point can affect the counts of more than two bins. Consequently, adding Laplace noise with scale \(2/\varepsilon\) may not be sufficient to guarantee \(\varepsilon\)-differential privacy.}
\end{tcolorbox}
{\sffamily\small\textbf{Model Response}~{\color{gray}(gpt-5-high)}}\par\vspace{2pt}
\begin{lstlisting}
Let I1,...,In be fixed (data-independent) intervals, possibly overlapping. Let h(D) in R^n be the histogram vector with coordinates hi(D) = #{x in D : x in Ii}. For replace-one neighbors D and D' that differ by changing one record from x to y, the change in the histogram is
h(D) - h(D') = (1{xinIi} - 1{yinIi})i=1..n,
so
||h(D) - h(D')||1 = sumi |1{xinIi} - 1{yinIi}| = |Sx Delta Sy|,
where Sx = {i : x in Ii} and Delta denotes symmetric difference.

Let c = maxz |Sz| be the maximum number of bins any single point belongs to (the overlap multiplicity). Then
||h(D) - h(D')||1 <= |Sx| + |Sy| <= 2c,
and this bound can be tight. For disjoint bins, c = 1 and the L1 sensitivity under replace-one is Delta1 = 2. For overlapping bins with c > 1, the global sensitivity Delta1 can exceed 2 (e.g., if some x lies in r > 1 bins and y in none, then ||h(D) - h(D')||1 = r).

The Laplace mechanism that adds i.i.d. Lap(b) noise per coordinate achieves epsilon' = Delta1/b differential privacy. With b = 2/epsilon, the achieved privacy is epsilon' = (epsilon Delta1)/2. If Delta1 > 2 (which occurs whenever c > 1 in the worst case), then epsilon' > epsilon, so the mechanism is not epsilon-differentially private.

A simple counterexample: take three identical intervals I1 = I2 = I3 = [-1,1]. Replacing a single record at x = 0 with y = 10 changes three coordinates by 1, so Delta1 = 3. Adding Lap(2/epsilon) yields epsilon' = 3epsilon/2 > epsilon, violating epsilon-DP.

Therefore, adding Lap(2/epsilon) noise per bin guarantees epsilon-DP only if the bins are disjoint (c = 1). In general with overlapping bins, it does not.

\boxed{\text{no}}
\end{lstlisting}

{\sffamily\small\textbf{Judge Assessment}~{\color{gray}(gpt-5.4)}}\par\vspace{2pt}
\begin{tcolorbox}[colback=catminorbg,colframe=catminor,boxrule=0.5pt,arc=2pt,left=8pt,right=8pt,top=5pt,bottom=5pt]
{\color{catcorrect}\textbf{Reasoning correct}}\hfill\textbf{Category:}~{\color{catminor}\textbf{Minor Gap}}\\[3pt]
\textbf{Summary:}~The response correctly argues via L1 sensitivity and gives a valid counterexample showing Lap(2/epsilon) can fail for overlapping bins, though it slightly overstates that disjointness is necessary in all cases.\\[3pt]
\textbf{Detail:}~The main reasoning is sound: for replace-one neighbors, the histogram change is the symmetric difference of bin-membership vectors, so overlapping intervals can make the L1 sensitivity exceed 2. Since the Laplace mechanism with coordinatewise Lap(2/epsilon) only guarantees epsilon-DP when the global L1 sensitivity is at most 2, the explicit example with three identical intervals gives sensitivity 3 and therefore violates epsilon-DP. The only imprecision is the statement that the mechanism guarantees epsilon-DP only if bins are disjoint; overlapping bins can still have sensitivity at most 2 in some special cases. But for the problem's yes/no claim, the counterexample is enough, so the conclusion is correct.
\end{tcolorbox}
\vspace{4pt}\noindent\rule{\linewidth}{0.4pt}

\vspace{6pt}
\begin{tcolorbox}[colback=catcorrect,colframe=catcorrect,boxrule=0pt,arc=3pt,left=6pt,right=6pt,top=5pt,bottom=5pt]\color{white}\begin{tabular}{@{}p{1.1cm}p{2.4cm}p{2.4cm}p{7.4cm}@{}}\textbf{Q\,96} & GT:~\textbf{No (0)} & Pred:~\textbf{No (0)} & \textbf{Correct}~~{\color{catcorrect}\checkmark\ correct}\end{tabular}\end{tcolorbox}
\begin{tcolorbox}[colback=gray!7,colframe=gray!35,boxrule=0.5pt,arc=2pt,left=8pt,right=8pt,top=5pt,bottom=5pt]\noindent{\small\color{gray}\textit{Accounting and Composition $\triangleright$ Parallel Composition}}\\[4pt]
We aim to release a histogram whose bins are $n$ disjoint, data-independent intervals. 
If we add independent noise drawn from $\mathrm{Lap}(1/\varepsilon)$ to each bin, 
does the resulting mechanism satisfy $\varepsilon$-differential privacy under the replace-one neighboring relationship?
\end{tcolorbox}
\vspace{2pt}
{\sffamily\small\textbf{Comments from metadata}}\par\vspace{2pt}
\begin{tcolorbox}[colback=blue!4,colframe=blue!30,boxrule=0.5pt,arc=2pt,left=8pt,right=8pt,top=5pt,bottom=5pt]
{\small The sensitivity for releasing the histogram under the replace-one neighboring relation is 2. Therefore, adding independent Laplace noise with scale $\varepsilon$ to each histogram bin only guarantees $2\varepsilon$-differential privacy.}
\end{tcolorbox}
{\sffamily\small\textbf{Model Response}~{\color{gray}(gpt-5-high)}}\par\vspace{2pt}
\begin{lstlisting}
Let h(D) in R^n be the histogram counts for n fixed, disjoint bins. Under the replace-one neighboring relationship, a single record can move from one bin to another, so exactly one bin decreases by 1 and another increases by 1. Hence the L1-sensitivity of the vector-valued query is
Delta1(h) = max_{D~D'} ||h(D) - h(D')||_1 = 2.

The vector Laplace mechanism that adds independent Lap(b) to each coordinate satisfies epsilon-DP when b = Delta1/epsilon. With b = 1/epsilon, the joint density of the mechanism is
p(y | D) propto exp(-epsilon ||y - h(D)||_1),
and for neighboring D, D':
p(y | D) / p(y | D') <= exp(epsilon ||h(D) - h(D')||_1) <= exp(2epsilon).

Thus the mechanism is 2epsilon-DP (not epsilon-DP) under replace-one adjacency. To achieve epsilon-DP under replace-one, one would need Lap(2/epsilon) noise per bin.

\boxed{\text{no}}
\end{lstlisting}

{\sffamily\small\textbf{Judge Assessment}~{\color{gray}(gpt-5.4)}}\par\vspace{2pt}
\begin{tcolorbox}[colback=catcorrectbg,colframe=catcorrect,boxrule=0.5pt,arc=2pt,left=8pt,right=8pt,top=5pt,bottom=5pt]
{\color{catcorrect}\textbf{Reasoning correct}}\hfill\textbf{Category:}~{\color{catcorrect}\textbf{Correct}}\\[3pt]
\textbf{Summary:}~The response correctly identifies that under replace-one adjacency the histogram has L1 sensitivity 2, so Laplace noise with scale 1/epsilon per bin yields 2epsilon-DP, not epsilon-DP.\\[3pt]
\textbf{Detail:}~The argument correctly notes that replacing one record can decrement one bin and increment another, changing the histogram by an L1 distance of 2. It then applies the standard Laplace mechanism analysis for vector-valued queries with independent Laplace noise, showing the privacy loss is bounded by exp(epsilon*2)=exp(2epsilon) when the scale is 1/epsilon. The conclusion that the mechanism is not epsilon-DP under replace-one adjacency, and would need scale 2/epsilon to achieve epsilon-DP, is mathematically sound.
\end{tcolorbox}
\vspace{4pt}\noindent\rule{\linewidth}{0.4pt}

\vspace{6pt}
\begin{tcolorbox}[colback=catfatal,colframe=catfatal,boxrule=0pt,arc=3pt,left=6pt,right=6pt,top=5pt,bottom=5pt]\color{white}\begin{tabular}{@{}p{1.1cm}p{2.4cm}p{2.4cm}p{7.4cm}@{}}\textbf{Q\,106} & GT:~\textbf{No (0)} & Pred:~\textbf{Yes (1)} & \textbf{Fatal Error}~~{\color{catfatal}\(\times\)\ wrong}\end{tabular}\end{tcolorbox}
\begin{tcolorbox}[colback=gray!7,colframe=gray!35,boxrule=0.5pt,arc=2pt,left=8pt,right=8pt,top=5pt,bottom=5pt]\noindent{\small\color{gray}\textit{Accounting and Composition $\triangleright$ Parallel Composition}}\\[4pt]
Consider the following procedure: $\mathcal{M} := \{\mathcal{M}_i(X_i)\}_{i=1}^n$, where $X_i \cap X_{i+1} = \emptyset$ for all $i\in\{1,\ldots,n-1\}$, and their union covers the entire dataset, i.e., $\bigcup_{i=1}^n X_i = X$. Furthermore, the partition is data-independent. 
Suppose each mechanism $\mathcal{M}_i$ satisfies $(\varepsilon_i, \delta_i)$-differential privacy.  
Does the combined mechanism $\mathcal{M}$ satisfy $(\max_{i \in [n]}\varepsilon_i,\max_{i \in [n]} \delta_i)$-differential privacy?
\end{tcolorbox}
\vspace{2pt}
{\sffamily\small\textbf{Comments from metadata}}\par\vspace{2pt}
\begin{tcolorbox}[colback=blue!4,colframe=blue!30,boxrule=0.5pt,arc=2pt,left=8pt,right=8pt,top=5pt,bottom=5pt]
{\small Consider the following counter example: let $X_i = X$ when $i$ is odd and $X_i = \emptyset$ when $i$ is even, then $\mathcal{M}$ satisfies \(\mathcal{M}\) satisfies $\Big(\sum\limits_{i \in [n]\text{ and i is odd}}\varepsilon_i, \sum\limits_{i \in [n]\text{ and i is odd}}\delta_i\Big)$-DP}
\end{tcolorbox}
{\sffamily\small\textbf{Model Response}~{\color{gray}(gpt-5-high)}}\par\vspace{2pt}
\begin{lstlisting}
Yes. This is the parallel composition property of differential privacy.

Explanation/derivation:
- Because the partition {X_i} is disjoint and data-independent, for any neighboring datasets X and X' that differ in exactly one individual's record, there exists a unique block j such that X_j != X'_j and X_i = X'_i for all i != j.
- For i != j the inputs to M_i are identical on X and X', hence the distributions of M_i(X_i) are identical across the two datasets and contribute no privacy loss.
- The joint release M(X) = (M_1(X_1), ..., M_n(X_n)) can be viewed as post-processing of M_j(X_j) together with side information and randomness that do not depend on the changed record (namely, the other blocks X_{-j}, which are the same across X and X', and any randomness). Differential privacy is preserved under post-processing and under adding randomness independent of the changed record.
- Therefore, for any event S, P[M(X) in S] <= e^{epsilon_j} P[M(X') in S] + delta_j. Maximizing over the possible affected block j gives (max_i epsilon_i, max_i delta_i).

Final answer:
\boxed{yes}
\end{lstlisting}

{\sffamily\small\textbf{Judge Assessment}~{\color{gray}(gpt-5.4)}}\par\vspace{2pt}
\begin{tcolorbox}[colback=catfatalbg,colframe=catfatal,boxrule=0.5pt,arc=2pt,left=8pt,right=8pt,top=5pt,bottom=5pt]
{\color{catfatal}\textbf{Reasoning incorrect}}\hfill\textbf{Category:}~{\color{catfatal}\textbf{Fatal Error}}\\[3pt]
\textbf{Summary:}~The response incorrectly applies parallel composition by assuming the \textbackslash{}(X\_i\textbackslash{}) form a fully disjoint partition, which is stronger than the stated condition.\\[3pt]
\textbf{Detail:}~The key step in the response--that a neighboring change affects a unique block \textbackslash{}(j\textbackslash{})--does not follow from the problem statement, which only requires \textbackslash{}(X\_i \textbackslash{}cap X\_\{i+1\} = \textbackslash{}emptyset\textbackslash{}), not pairwise disjointness for all \textbackslash{}(i \textbackslash{}neq j\textbackslash{}). A single individual's record may therefore appear in multiple non-adjacent subsets, so several mechanisms can incur privacy loss and the losses can compose additively rather than via a max. Thus the claimed \textbackslash{}((\textbackslash{}max\_i \textbackslash{}varepsilon\_i, \textbackslash{}max\_i \textbackslash{}delta\_i)\textbackslash{})-DP guarantee is not justified for the stated setup, and the final conclusion "yes" is incorrect.
\end{tcolorbox}

\section{Paraphrasing experiment of \catwo{}}\label{apx:para}
In this section we evaluate: Do LLMs rely on memorization for achieving high accuracy of positive instances? Standard DP textbooks and research papers may have appeared in model pretraining data, and many positive instances in our benchmark are faithful re-statements of results with the same notations established in the literature. This raises the concern that LLMs might rely on superficial pattern matching -- simply answering “yes” when a question resembles previously seen material. To probe this possibility, we conduct an ablation study in which all positive instances in \catwo{} are paraphrased and re-evaluated using the two best-performing models, \gpthigh{} and \gemhigh{}. As shown in Table~\ref{tab:para}, accuracy changes only marginally, suggesting that performance is not driven solely by memorization but also the understanding. 

\begin{table}[H]
\centering
\caption{Average accuracy on paraphrased positive questions, results are reported in format Mean Accuracy$\pm$ standard deviation}
\label{tab:para}
\fontsize{8}{10}\selectfont
\begin{tabular}{lcc}
\toprule
Model & After Paraphrase & Before Paraphrase \\
\midrule
Gemini-3-pro & $0.810 \pm 0.024$ & $0.805\pm0.039$ \\
GPT-5-High & $0.771 \pm 0.055$ & $0.771\pm 0.027$ \\
\bottomrule
\end{tabular}
\end{table}

\subsection{Paraphrasing Set-up}

For each positive question in \catwo{}, we use GPT-5 to generate a paraphrased version. We then let a human expert verify that the paraphrase preserves the original question’s meaning and fixes any LaTeX compilation errors when necessary. The system prompt used for paraphrasing is provided below.

\begin{tcolorbox}[
    colback=gray!5,
    colframe=cyan!75!black,
    title=System prompt for paraphrasing question,
    fonttitle=\bfseries
]
{
You are given a technical description of a differential privacy mechanism.\\
\\
Task:\\
- paraphrasing, change notation and algorithm style, but don't change name of algorithm\\
- Change the structure of the algorithm (change the order of the steps, represent the operations in a different way)\\

Constraints:\\
- Preserve exact mathematical meaning.\\
- Change all variable names, symbols, and function identifiers.\\
- Rewrite the algorithm using a different presentation style.\\
- Do NOT change the algorithm name.\\
- ALL mathematical symbols, formulas, and equations MUST be written in valid LaTeX.\\
- Do NOT introduce non-LaTeX math notation.\\
- Do NOT add explanations, commentary, or assumptions.\\
- Output ONLY the paraphrased LaTeX content.
}
\end{tcolorbox}

\subsection{Example of Paraphrased Questions}

\begin{figure}[H]    
  \begin{minipage}{0.42\textwidth}
    \centering
    \includegraphics[width=\linewidth]{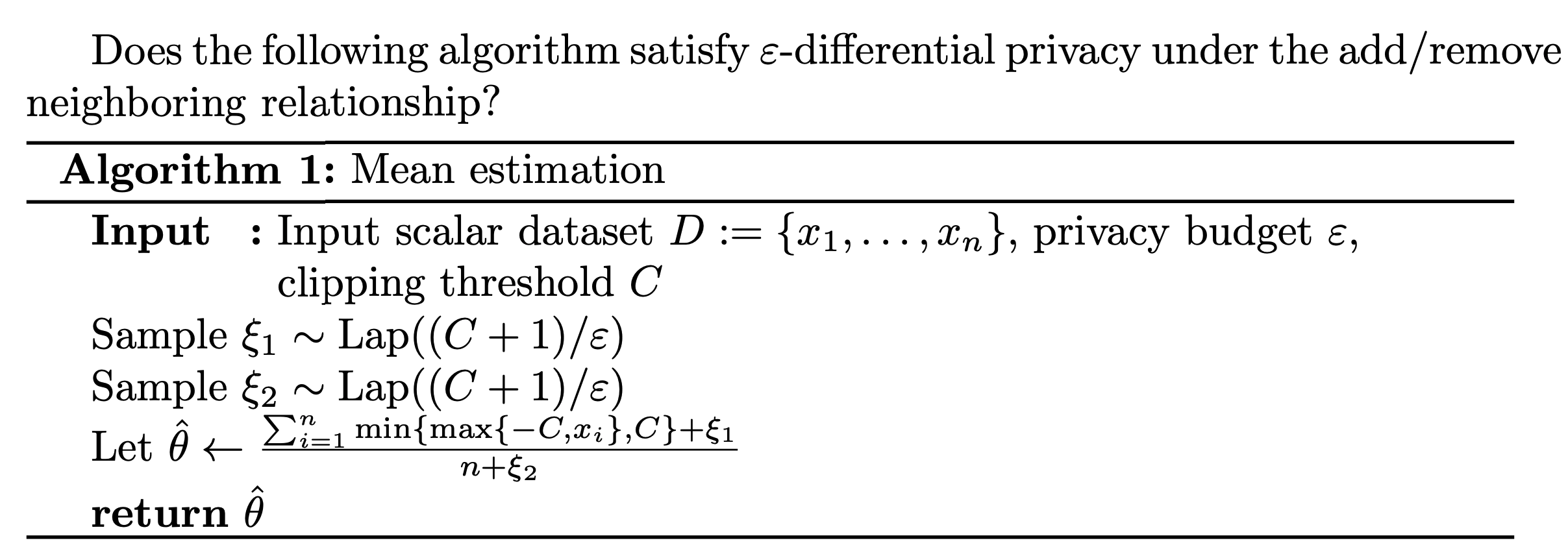}
    \subcaption{Question 72: Original}\label{fig:para_c}
  \end{minipage}
  \hfill
  \begin{minipage}{0.42\textwidth}
    \centering
    \includegraphics[width=\linewidth]{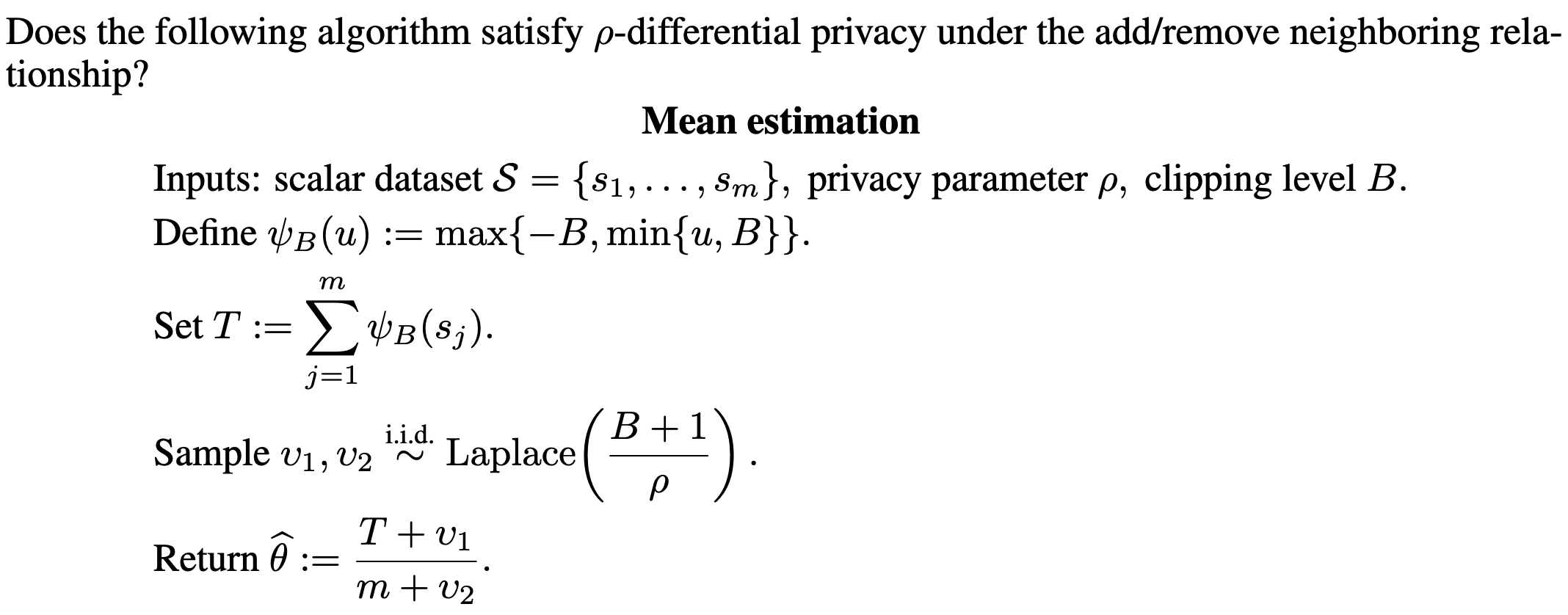}
    \subcaption{Question 72: Paraphrased}\label{fig:para_d}
  \end{minipage}
  
  \vspace{0.5cm} %

  \centering
  \begin{minipage}{0.42\textwidth}
    \centering
    \includegraphics[width=\linewidth]{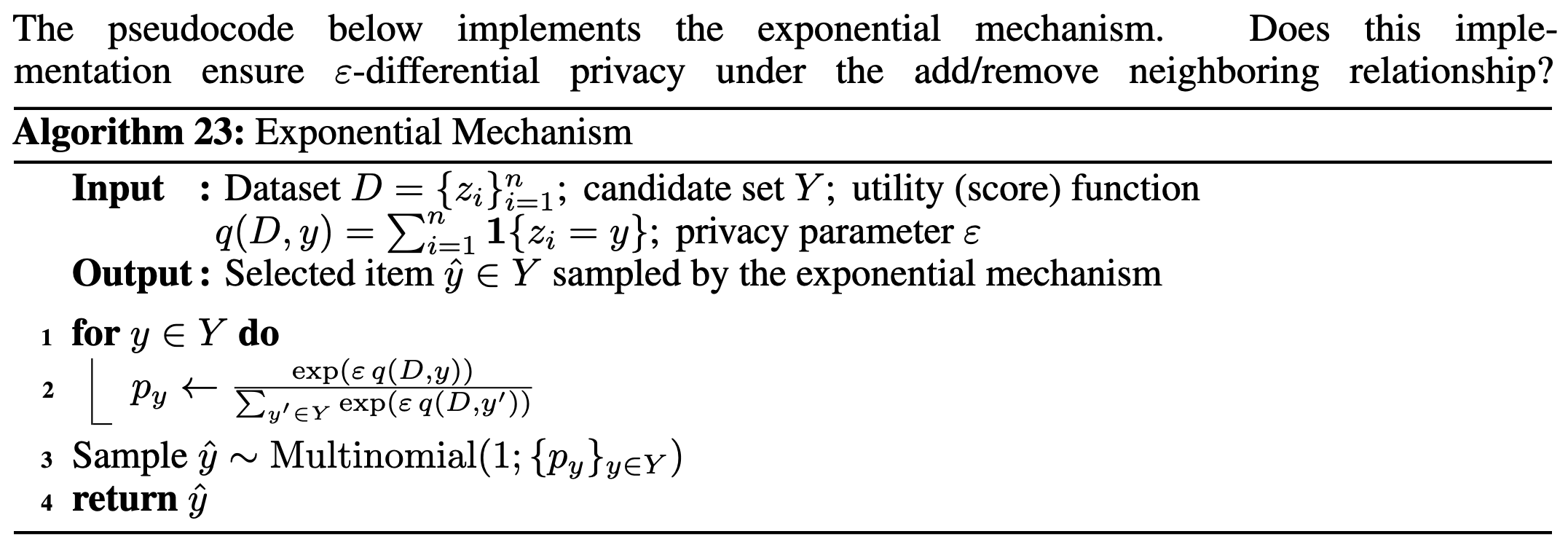}
    \subcaption{Question 18: Original}\label{fig:para_a}
  \end{minipage}
  \hfill
  \begin{minipage}{0.42\textwidth}
    \centering
    \includegraphics[width=\linewidth]{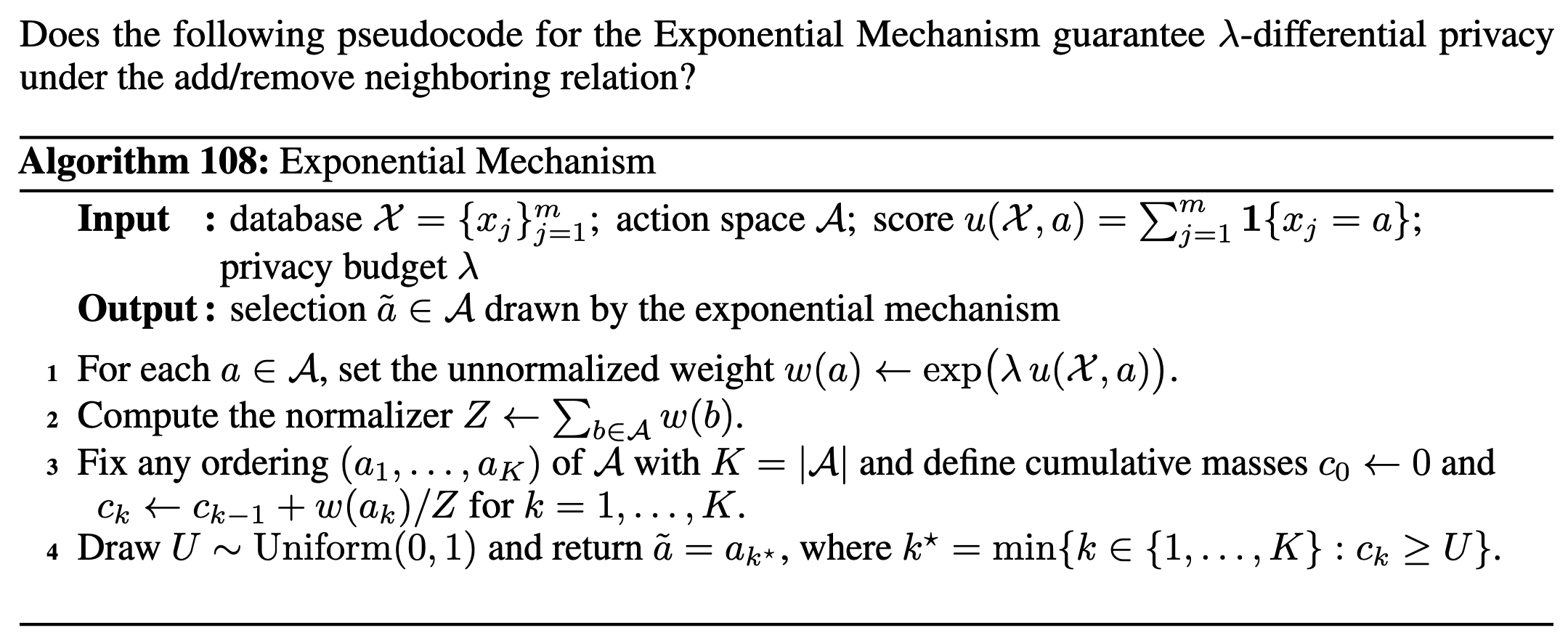}
    \subcaption{Question 18: Paraphrased}\label{fig:para_b}
  \end{minipage}

  \caption{Examples of paraphrased questions.}
  \label{fig:para_experiment}
\end{figure}

We identified some interesting patterns in the paraphrased questions. In some cases, the paraphrased versions exhibit only symbolic differences or superficial changes in structure, as shown in Figure~\ref{fig:para_experiment}, panels (a) and (a). However, the structure of the paraphrased algorithm can also change in a more substantive way. As shown in Figure~\ref{fig:para_experiment} (c) and (d), the high-level description of the exponential mechanism (Lines 2-3 of Figure~\ref{fig:para_a}) has been paraphrased to inverse CDF sampling (Lines 3-4 of Figure~\ref{fig:para_b}).

\end{document}